\documentclass[twocolumn]{svjour3}          

\usepackage[a4paper, margin=2cm]{geometry}
\usepackage{xcolor}

\smartqed  
\usepackage{graphicx}
\usepackage{booktabs}
\usepackage{amssymb}
\usepackage{amsmath}
\usepackage{subfigure}
\usepackage{natbib}
\usepackage[]{siunitx}
\usepackage[ruled,vlined]{algorithm2e}
\usepackage{hyperref}
\RequirePackage{fix-cm}
\newcommand{\vc}[1]     {\ensuremath{\boldsymbol{#1}}}
\newcommand{\vx}        {\vc{x}}
\newcommand{\pars}      {\vc{\beta}}
\newcommand{\rules}      {\vc{r}}
\newcommand{\hyper}      {\vc{\eta}}

\newcommand{\vy}        {\vc{y}}

\newcommand{\vf}        {\vc{f}}

\journalname{Preprint}

\begin{document}

\title{Rule-based Bayesian regression
}

\titlerunning{Rule-based regression}        

\author{Themistoklis Botsas,$^{a}$
	Lachlan R. Mason,$^{a,b}$ Indranil Pan$^{a,b}$ \\
	\newline
	$^{a}$The Alan Turing Institute \\
	$^{b}$Imperial College London \\
	}

\authorrunning{Botsas, Mason, Pan} 


\date{Received: date / Accepted: date}

\maketitle

\begin{abstract}
We introduce a novel rule-based approach for handling regression problems. The new methodology carries elements from two frameworks: (i) it provides information about the uncertainty of the parameters of interest using Bayesian inference, and (ii) it allows the incorporation of expert knowledge through rule-based systems. The blending of those two different frameworks can be particularly beneficial for various domains (e.g. engineering), where, even though the significance of uncertainty quantification motivates a Bayesian approach, there is no simple way to incorporate researcher intuition into the model. We validate our models by applying them to synthetic applications: \textcolor{black}{a simple linear regression problem and two more complex structures based on partial differential equations, and we illustrate their use through two cases derived from real data}. Finally, we review the advantages of our methodology, which include the simplicity of the implementation, the uncertainty reduction due to the added information and, in some occasions, the derivation of better point predictions, and we \textcolor{black}{outline} limitations, mainly from the computational complexity perspective, such as the difficulty in choosing an appropriate algorithm and the added computational burden.
\end{abstract}


\section{Introduction}\label{Introduction}

Expert knowledge elicitation and their incorporation in statistical models play an important role in statistical inference and evidence based decision making \citep{o2019expert}. Most methods, however, look at ways of expressing knowledge about an uncertain quantity in the form of a (subjective) probability distribution. In this paper we investigate the suitability of rule based systems as a framework to integrate expert knowledge into statistical models. In general, rule based systems like decision trees are used as statistical models themselves and have seen wide applicability in multiple applied domains \citep{breiman1984classification}. Regression and classification trees commonly used in such learning methods partition the input feature space using hierarchical rules to map onto the target variable. The interpretable nature of these tree-based models has made them gain renewed traction in the quest for explainable machine learning models \citep{lundberg2020local}. In addition to the traditional variants of decision-tree learning that provide point predictions, probabilistic counterparts like Bayesian additive regression trees (BARTs) \citep{chipman2010bart} and Mondrian forests \citep{lakshminarayanan2016mondrian} have also been developed. The rules in these systems are algorithmically generated through recursive partitioning or other allied algorithms. Such algorithmically generated rules may give high predictive accuracy, but they are often non-intuitive. Moreover, in traditional decision-tree based learning paradigms there are no provisions for explicit incorporation of prior expert rule bases which are often readily available in multiple engineering application domains.

Standard Bayesian regression techniques that use hierarchical models \citep{gelman2013bayesian}, Gaussian processes (GPs) \citep{rasmussen2003gaussian} or splines \citep{de1978practical}, offer alternative statistical approaches whereby incorporation of expert opinion is possible in the form of prior distributions of model (hyper) parameters. However, expert elicitation methods generally do not incorporate prior knowledge in the form of rules. \textcolor{black}{Examples of such expert knowledge include constraints on outputs based on the nature of the inputs (i.e. with high values of the input  variable \textit{a} we expect high values of output \textit{c}, or with input \textit{a} smaller than input  \textit{b} we expect negative output \textit{c}). Even though there are simple ways in which the above sentences can be demonstrated with rules, there is rarely a direct equivalence with a Bayesian prior.} A previous study conducted by \citet{pan2017fuzzy} focused on the importance of commonly available expert knowledge in engineering domains which cannot be effectively incorporated in traditional Bayesian modelling techniques.

In Section~\ref{Rule-systems} we present the general framework of rule-based systems and we demonstrate how expert knowledge is encoded in the form of IF–THEN logic-based rules, but can also be composed into more complex rule bases with logic operations (AND, OR, NOT). In Section \ref{Methodology} we build upon the study of \citet{pan2017fuzzy} to provide a pure probabilistic framework for expressing expert-elicited rule bases in a Bayesian context: the framework can be used in conjunction with standard statistical regression methods. Such an approach can be seen as leveraging the best of both worlds, i.e.~using interpretable rule-based methods and taking advantage of the flexibility (and consequently high predictive accuracy) of data-driven regression methods. In Section \ref{Applications} we develop intuition with simple case studies \textcolor{black}{and apply it to the case of more complicated spatio–temporal differential equations, as well as actual field problems, in order to show distinctive advantages of the proposed framework}. In Section \ref{discussion} we address the shortcomings due to computational complications and highlight relevant future work needed in this area. Finally, in Section \ref{conclusion} we summarise the main takeaways from our work, focusing on the flexibility and simplicity of the new methodology and discuss the focus of our future research.

\section{Rule-based systems}\label{Rule-systems}

Many popular data-driven algorithms (e.g.~linear regression, spline regression) can easily be extended to their full Bayesian counterparts, while others incorporate a form of embedded uncertainty quantification (e.g.~GPs). In real-world applications, these Bayesian variations are widely used, not only because they provide a principled way to quantify system uncertainty, but also because they allow for inclusion of the domain expertise in the form of conventional informative priors. Unfortunately, converting expert knowledge into a prior distribution can be exceptionally challenging, since it is not generally trivial to associate external knowledge with data-driven model parameters.

As a result, there is not always an obvious approach to incorporate insights and intuition (such as the structure of the outputs, given the structure of the inputs) into the prior distribution, and, therefore, researchers need to take alternative actions to integrate this form of knowledge into their systems. One such approach is the use of \textit{rule-based systems}.

Our \emph{rule-based} definition includes systems that incorporate knowledge in the form of human-crafted rule base $R_k$, which can be expressed as:
\begin{equation}
    \delta_k R_k : \text{ if } A_1^k \oplus A_2^k \oplus \dots \oplus A_{m}^k \text{ then } C_k,
\end{equation}
where $\delta_k$ is a dichotomous variable indicating the inclusion of the $k$th rule in the system; $A^k_{i}$, $i \in {1, 2,\dots, m }$, is the value of the $i$th antecedent attribute (cause) in the $k$th rule; $l$ is the number of antecedent attributes used in the $k$th rule; $C_k$ is the consequent (effect) in the $k$th rule; and $\oplus \in \{\lor,\land\}$ represents the set of connectives (OR, AND operations) in the rules.

For rule-based Bayesian regression, we include a logical-operator-based (AND, OR) combination of all the rules to give rise to a composite rule base: i.e., $\beta_k = 1,\ \forall\ k$, and we use the quantity:
\begin{equation*}
    R_\text{comp} :=  R_1 \oplus R_2 \oplus \dots \oplus R_n.
\end{equation*}
Finally, in our context, the antecedent attributes are functions of the inputs (e.g.~summary statistics) and the consequent is a function of the outputs. Concrete examples are described in Section~\ref{Applications}.

\section{Methodology}\label{Methodology}

We now aim to incorporate the rule-based system from Section \ref{Rule-systems} into the standard Bayesian framework.

\subsection{Rule-based Bayesian context}\label{sec:rule_bayes}

The posterior density in a typical Bayesian context, is provided by Bayes' theorem:
\begin{equation}\label{Bayes}
    p(\pars | \vx) = \frac{p(\vx | \pars) p(\pars)}{p(\vx)},
\end{equation}
where $\vx$ are the data, $\pars$ are the model parameters, $p(\vx | \pars)$ is the likelihood (a measure of goodness of fit of the model to the data), and $p(\pars)$ is the prior density accounting for knowledge about the system before data are taken into account. The model evidence, $p(\vx)$, otherwise known as the marginal likelihood, acts as a normalising constant and is also the probability of obtaining the observed data with the effect of the parameters marginalized. Finally, $p(\pars | \vx)$ is the posterior density; it is the main quantity of interest and it reflects our updated knowledge about the model parameters after we include the information from the observed data.

\begin{figure*}
    \centering
    \includegraphics[width=\textwidth]{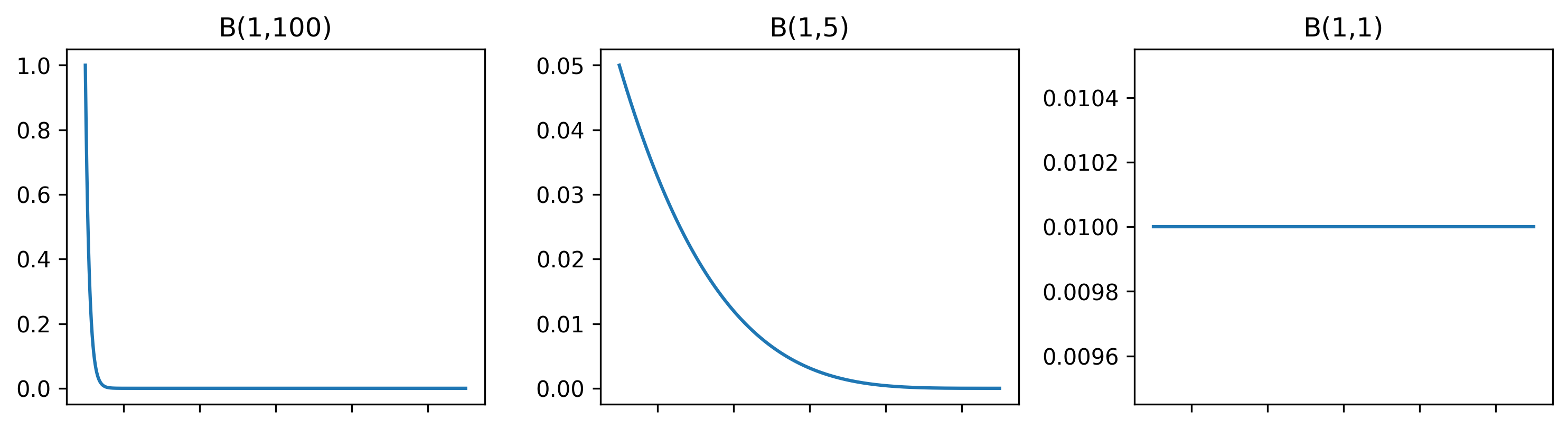}\label{fig:Beta}\\
    \caption{\textcolor{black}{Indicative cases of the Beta distribution that can be used for the density of $\rules | \pars$. $\operatorname{Beta}(1,100)$ (left) corresponds to a strict rule, $\operatorname{Beta}(1,5)$ (middle) to a non-strict rule and $\operatorname{Beta}(1,1)$ (right) to equivalent to a non-rule system.}}
    \label{fig:Betadist}
\end{figure*}

Since the model evidence is generally difficult to compute, we, instead, resort to the proportionality variation of Bayes' theorem:
\begin{equation}\label{Bayes_prop}
    p(\pars | \vx) \propto p(\vx | \pars) p(\pars),
\end{equation}
which can be calculated using specific algorithmic classes, such as Markov Chain Monte Carlo (MCMC) methods.

In order to combine the rule-based approach within the Bayesian formalism, we modify the prior distribution. Equation \ref{Bayes_prop} then becomes:
\begin{equation}\label{Bayes_rules1}
    p(\pars | \vx, \rules) \propto p(\vx | \pars) p(\pars , \rules),
\end{equation}
where $\rules$ is a random variable associated with the rule base, and it is going to be explained in detail later. The joint distribution $p(\pars , \rules)$ reflects our prior knowledge from the two sources, namely (i) the conventional prior information regarding the model parameters $\pars$ and (ii) the expert information from the rules, \textcolor{black}{associated with the variable} $\rules$. In order to compute this quantity, we use the probability chain rule:
\begin{equation*}
    p(\pars , \rules) = p(\rules | \pars) p(\pars).
\end{equation*}
The conditional probability $p(\rules | \pars)$ can be perceived as the probability of obtaining the variable $\rules$ given the proposed model parameters $\pars$. Its effect is similar to that of hyperparameters in a conventional Bayesian hierarchical context.

In practice, we pre-define a set or discretisation of \textit{rule-input} values that correspond to the rule antecedents. We obtain the number of rule-input points that violate the rule consequents (for a set of parameters $\pars$) and we take its ratio to the number of all rule inputs. This ratio corresponds to $\rules | \pars$. \textcolor{black}{It is also worth noting that in a set-up with many different rules, we can combine different sizes of discretisation. In that sense, the method favors the rules with finer resolution more than the ones with coarse resolution. Alternatively, $\rules | \pars$ can be extended to a single distribution for each rule, which allows for more flexibility for the model, since different levels of confidence can be considered for individual rules.}

In our simulations we use a beta distribution; $\rules | \pars \sim \operatorname{Beta}(a,b)$, where different values of the parameters $a$ and $b$ account for different levels of confidence in the rule base. \textcolor{black}{Technically, any distribution can be used for $\rules | \pars$, but the form of a beta distribution is particularly useful, since, for specific values of its parameters, it is very intuitive to adjust its parameters based on the rules' confidence. In Figure \ref{fig:Betadist} we show three cases of the Beta distribution with different shape parameters. The first case, where $\rules | \pars \sim \operatorname{Beta}(1,100)$ corresponds to a very strict penalty. The ratio mentioned above has a very high probability when none of the rule-input points violate the rules (left plot) and is almost zero everywhere else. The second case, where, $\rules | \pars \sim \operatorname{Beta}(1,5)$, can be related to a non-strict rule. In this case (middle plot), the density tends to zero much slower, and, therefore, the magnitude of the penalty is much smaller when only few of the points violate the rule, while in cases where most points violate the rule we still observe probability close to zero. In the last case (right plot), where $\rules | \pars \sim \operatorname{Beta}(1,5)$ (which is also equivalent to a uniform distribution), the same penalty is added regardless of the amount of rule-input points that violate the rule. This case is identical to the standard, non-rule-based approach.}

\textcolor{black}{To sum up, there are two main components that construct the rule-based Bayesian regression: The discretisation and the distribution of $\rules | \pars$. The former determines how many points violate the rule, while the latter the magnitude of the penalty added to the likelihood based on these violations. Due to this penalty, some of the MCMC samples, that otherwise would be accepted, are instead, rejected. This has an effect on the shape of the posterior.}

Combining the two formulae, the (un-normalized) posterior becomes:
\begin{equation}\label{eq:rule_regr}
    p(\pars | \vx, \rules) \propto p(\vx | \pars) p(\rules | \pars) p(\pars).
\end{equation}

\textcolor{black}{Note that in Equation \ref{eq:rule_regr} $\rules$ does not depend on the data, which reinforces the idea of equivalence to a hyperparameter prior in a fully Bayesian context.}

We can extend the model, by using hyperparameters $\hyper$ for the rules, which could account either for the structure of the rules (antecedents and consequents) or the parameters of the Beta distribution ($a$ and $b$). With use of the probability chain rule and assuming that the priors of $\pars$ and $\hyper$ are independent, Equation \ref{eq:rule_regr} becomes:
%
\begin{equation}\label{eq:rule_regr_hyper}
    p(\pars | \vx, \rules, \hyper) \propto p(\vx | \pars) p(\rules | \pars, \hyper) p(\pars) p(\hyper).
\end{equation}

\subsection{Rule-based Gaussian process regression}\label{sub:theory_GPR}

In GP regression, the main quantity of interest is the marginal likelihood \citep{rasmussen2003gaussian}. For a set of inputs $\vx$, a set of outputs $\vy$, i.i.d. Gaussian noise $\epsilon$, and function value $\vf$, we obtain the marginal likelihood by integrating the product of a Gaussian likelihood $p(\vy|\vf, \vx)$ and the GP prior $p(\vf| \vx)$ with mean $m(\vx)$ and covariance kernel $k(\vx,\vx')$ over the function values $\vf$:
\begin{equation}\label{eq:marg_lik}
    p(\vy | \vx) = \int_{\vf} p(\vy|\vf, \vx)p(\vf | \vx)\, \mathrm{d}\vf,
\end{equation}
where
\begin{align*}
\vy &= \vf(\vx) + \boldsymbol{\epsilon},  \\
\boldsymbol{\epsilon} &\sim \mathcal{N}(0, \sigma_n^2)  \\
\vf(\vx) &\sim \operatorname{GP}(m(\vx), k(\vx,\vx')).
\end{align*}

Regarding predictions for new data $\vx_*$, the predictive distribution, which has a Gaussian form, is used:

\begin{equation}\label{eq:pred_dist}
    p(\vy_* | \vy) = \int p(\vy_* | \vf, \vx) p(\vf | \vy)\, \mathrm{d}\vf.
\end{equation}
\textcolor{black}{In what follows, we aim to derive the equivalent of a rule-based version of a Gaussian process marginal likelihood.}


Considering as $\vx$ and $\vy$ the data, $\vf$ the prior and $\rules$ a hyper-prior that depends on the data, the un-normalized posterior becomes:

\begin{align}\label{eq:gp_intro}
  p(\vf | \vy, \rules) &\propto p(\vy | \vf, \vx) p(\vf|\vx) p(\rules | \vf, \vy, \vx)\nonumber \\
  &\propto p(\rules, \vy, \vf | \vx).
\end{align}

\textcolor{black}{Note that in Equation \ref{eq:gp_intro}, $\rules$ depends on the data in contrast to Section \ref{sec:rule_bayes}. In practice, in Gaussian Process regression, when estimation for the covariance parameters is required, an approximation of the posterior, known as type II maximum likelihood \citep{rasmussen2003gaussian} is used. In order to get the equivalent density for the rule-based version, and since the model is no longer strictly Bayesian, we allow $\rules$ to be dependent on the data.}

Ideally, following the standard GP regression, we would integrate the RHS (right-hand side) of Equation \ref{eq:gp_intro} with respect to $\vf$, in order to derive a variation of the marginal likelihood in Equation \ref{eq:marg_lik}:

\begin{equation*}
    \int_{\vf} p(\rules, \vy, \vf |\vx) \, \mathrm{d}\vf = p(\rules, \vy| \vx).
\end{equation*}

Unfortunately, this is not possible in this case, since the term $p(\rules | \vf, \vy, \vx)$ violates the conjugacy of the GP model. As a substitute, we use a penalised version of the marginal likelihood, by constructing a penalty based on the prediction of a rule input. For this we employ 
\begin{equation*}
    \vy_r | \vy, \vf, \vx,
\end{equation*}
which is the prediction of the rule output for data $\vy$ and specific function values $\vf$. Equation \ref{eq:gp_intro} becomes:
\begin{equation}\label{eq:gp_intro2}
  \widehat{p}(\vf | \vy, \rules) \propto  p(\vy | \vf, \vx) p(\vf | \vx) p(\rules | \vy_r ).
\end{equation}
We calculate $\vy_r$ using summary statistics (e.g. the mean) of an equivalent version of the predictive distribution in Equation \ref{eq:pred_dist} for function values $\vf$. From Equation \ref{eq:gp_intro2}, we derive the intuition for the \emph{pseudo-marginal likelihood}:
\begin{equation}\label{eq:mod_marg_like}
    \widehat{p}(\vy,\rules | \vx) = p(\vy | \vx)p(\rules | \vy_r).
\end{equation}
Note that $\widehat{p}(\vy,\rules | \vx)$ is not a Gaussian distribution. In practice, we use an optimisation technique in order to compute the maximum a posteriori estimate (MAP), which corresponds to the mode of the posterior distribution. In that sense, the quantity $p(\rules | \vy_r )$ is equivalent to a penalty in a penalised maximum likelihood setting. The process to compute the pseudo-marginal likelihood in each step of the optimisation technique of choice is presented in Algorithm \ref{alg:pseudo-marg_like}.

\begin{algorithm}
\SetAlgoLined
\While{the optimisation algorithm has not converged}{
Compute $p(\vy | \vf, \vx) p(\vf | \vx)$ for specific function values $\vf$\;
Estimate the rule outputs $\vy_r$ from $\vy_r | \vy, \vf, \vx$ for some rule inputs $\vx_r$ and the function values $\vf$\;
Calculate the number of rule outputs that violate the rule and divide with the total number of rule inputs\;
Compute the rule penalty $p(\rules | \vy_r )$\;
Compute the product $p(\vy | \vf, \vx) p(\vf | \vx)p(\rules | \vy_r )$\;
Update the values $\vf$\;
}
 \caption{Compute the pseudo-marginal likelihood}\label{alg:pseudo-marg_like}
\end{algorithm}

As in Section \ref{sec:rule_bayes}, we are going to use a beta distribution; $\rules| \vy_r \sim \operatorname{Beta}(a,b)$, which accounts for the level of confidence in the rule base.

\section{Applications}\label{Applications}
We here illustrate the use of our methodology with the help of three synthetic \textcolor{black}{and two real world} applications. The first comprises a simple linear model. In the second, data are generated from a one-dimensional advection equation \citep{bar2019learning}, and we fit them with B-splines. In the third, we use data from a two-dimensional advection–diffusion equation \citep{Hoyer2020data} and we use a GP regression model. \textcolor{black}{We derive the data for the fourth from \citep{tufekci2014prediction}, and we use multivariate linear regression to predict the full load electrical power output of a combined cycle power plant. Finally, in the fifth, we use data from \citep{kaya2019predicting} in order to predict CO emissions from gas turbines, using a multivariate linear regression model. For the last two examples, the rules are derived from insights either directly from a field expert or from relevant literature}.
The results are produced using the PyMC3 Python package \citep{Salvatier2016} and the source code has been made available online.\footnote{\url{https://github.com/themisbo/Rule-based-Bayesian-regr}}  

\subsection{Linear regression}\label{sec:lin_reg}

In the following subsections, we first illustrate how the synthetic data were produced from a simple linear model and we present and compare the analyses for the standard Bayesian linear regression, the rule-based Bayesian linear regression with two different rule bases (one strict and one non-strict) and, finally, the rule-based Bayesian linear regression with hyperparameters.

For each case, we use a standard Metropolis–Hastings MCMC \citep{hastings1970monte}, with $4$ chains of \num{100500} iterations each, from which the first \num{500} are treated as burn-in. For the posterior plots we use thinning of \num{100}. In total, \num{4000} iterations are used for the results. For the intercept and slope priors, we use $\alpha \sim \mathcal{N}(0.5, 0.5^2)$ and $\beta \sim \mathcal{N}(0.5, 0.5^2)$ respectively.

\subsubsection{Data}

\begin{figure}
  \includegraphics[width=\columnwidth]{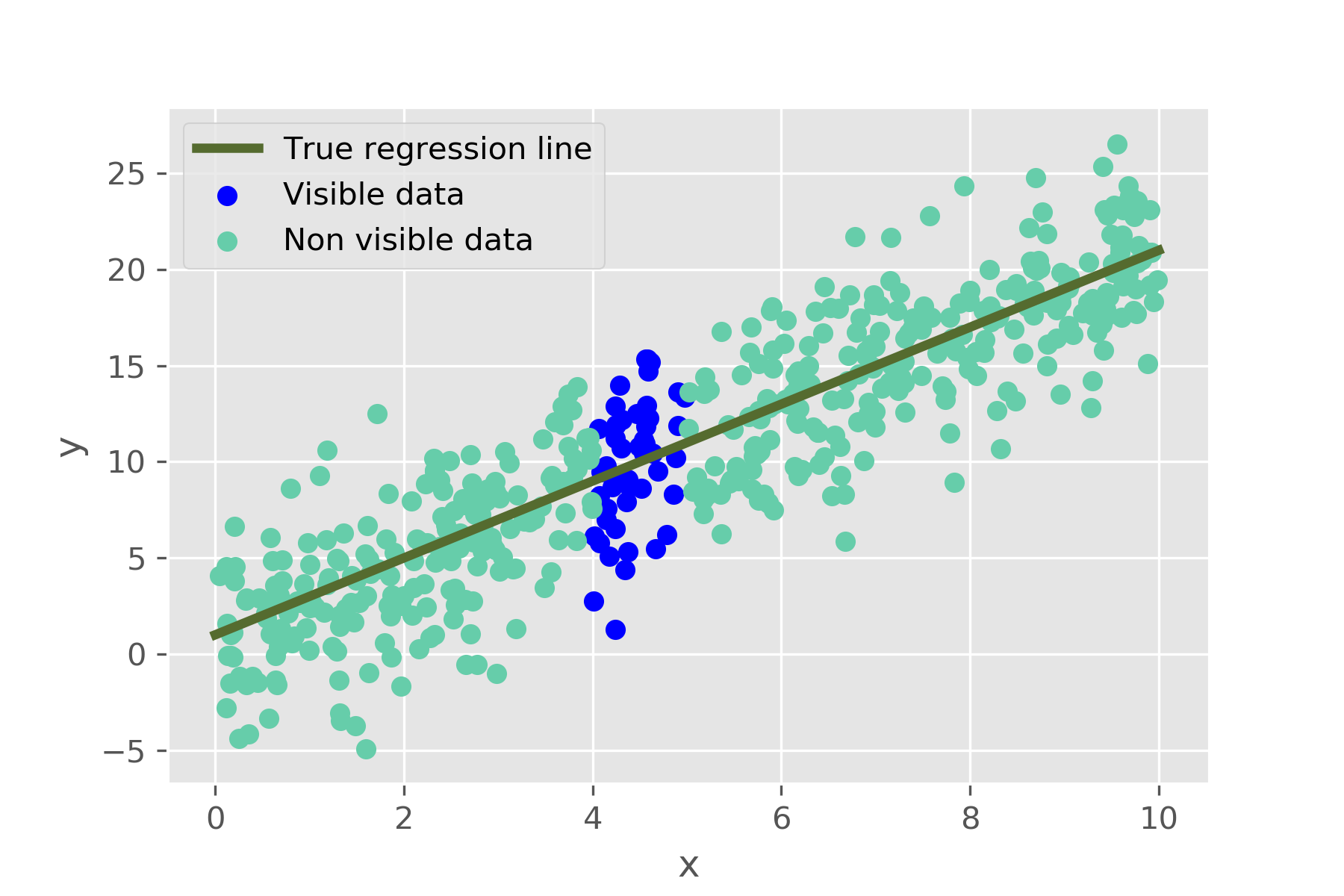}
\caption{Linear regression data. The dark blue points are used for the analysis.}
\label{fig:linreg_data}
\end{figure}

For this first application we produce synthetic linear data, from which we use only a small portion for the analysis. Our goal is to compensate for the lost information by adding information from a rule base and to examine changes in the predictive results and corresponding uncertainty. We initially sample $500$ random values within the interval $[0, 10]$. We produce the corresponding labels from the \textit{true} regression line $y = 1 + 2x + \epsilon$, where $\epsilon \sim N(0,3^2)$. From those points, we select only those within the sub-interval $[4,5]$ which leaves $49$ points for the final analysis step. The outcome is shown in Figure \ref{fig:linreg_data}.

\subsubsection{Bayesian linear regression (BLR)}

\begin{table*}
\caption{Posterior means $\mu$ and standard deviations $\sigma$ for the parameters $\alpha$ and $\beta$ for the true values (True), Bayesian linear regression (BLR), rule-based Bayesian linear regression with strict rules (RBLR-s), rule-based Bayesian linear regression with non-strict rules (RBLR-l) and rule-based Bayesian linear regression with hyperparameters (RBLR-h).}
\centering
\begin{tabular}{l cc cc cc cc cc}
    \toprule
             & \multicolumn{2}{c}{\textcolor{black}{True}} & \multicolumn{2}{c}{BLR} & \multicolumn{2}{c}{RBLR-s} & \multicolumn{2}{c}{RBLR-l} & \multicolumn{2}{c}{RBLR-h}                                         \\
    \cmidrule(lr){2-3} \cmidrule(lr){4-5} \cmidrule(lr){5-7} \cmidrule(lr){8-9} \cmidrule(lr){10-11}
    Metric   & \textcolor{black}{$\mu$}                   & \textcolor{black}{$\sigma$}           & $\mu$                   & $\sigma$                   & $\mu$                      & $\sigma$                   & $\mu$  & $\sigma$ & $\mu$  & $\sigma$ \\ \midrule
    $\alpha$ & \textcolor{black}{$1.00$}                  & \textcolor{black}{$1.73$}                     & $0.77$                  & $0.49$                     & $0.79$                     & $0.40$                     & $0.79$ & $0.40$   & $0.72$ & $0.38$   \\
    $\beta$  & \textcolor{black}{$2.00$}                  & \textcolor{black}{$1.73$}                     & $2.00$                  & $0.14$                     & $2.01$                     & $0.08$                     & $2.01$ & $0.09$   & $2.04$ & $0.08$   \\ \bottomrule
\end{tabular}
\label{tab:linreg_stats}
\end{table*}

\begin{figure}
  \includegraphics[width=\columnwidth]{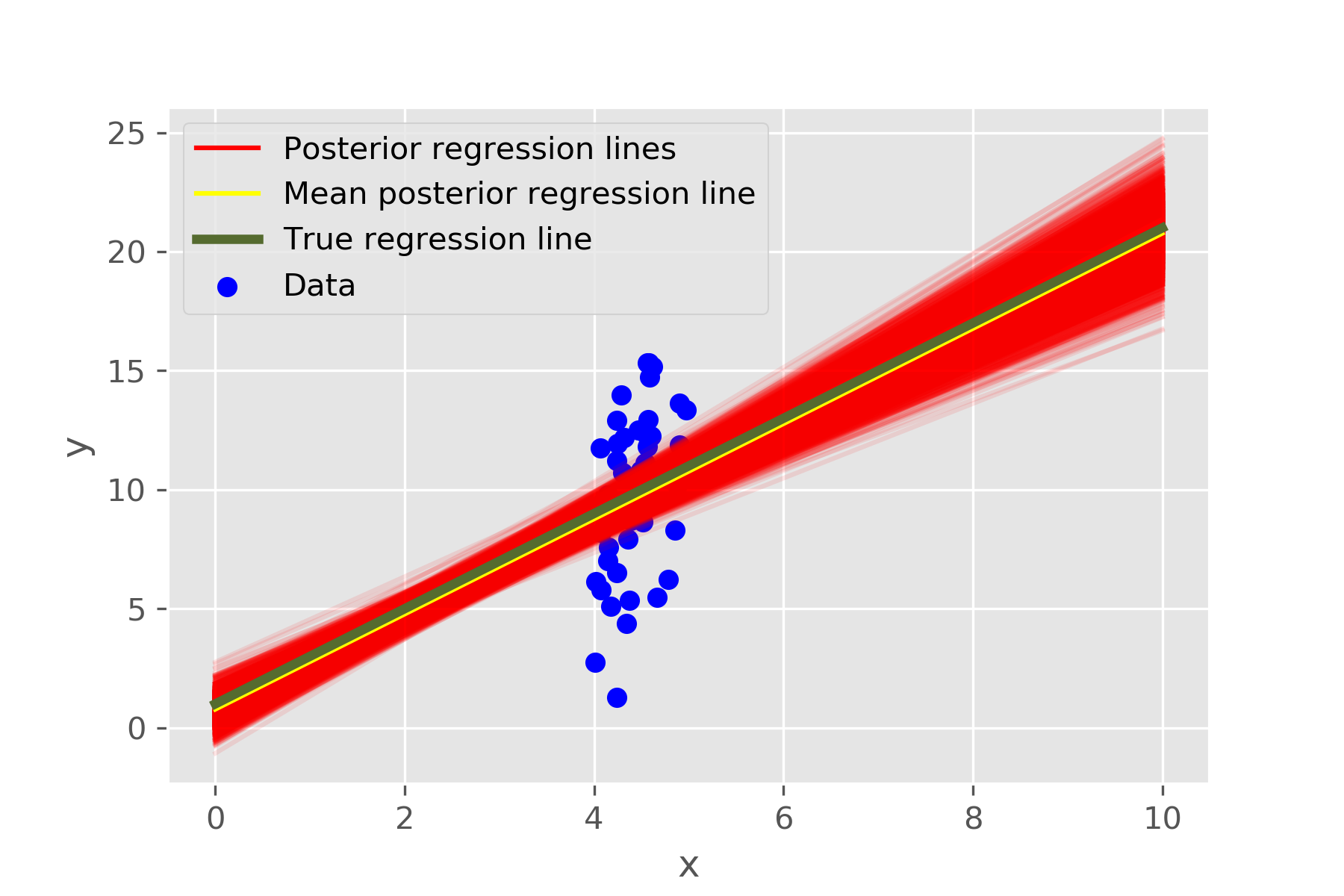}
\caption{Posterior regression lines. The red lines are realisations of (thinned) samples of the MCMC chain. The yellow line is the mean posterior regression line and the green line is the true regression line.}
\label{fig:linreg_post}
\end{figure}

The results for the linear regression model are shown in Figure \ref{fig:linreg_post}. The true regression line is very close to the mean posterior regression line. The absence of data, and consequently, lack of information in the left and right sections of the figure causes the uncertainty range to be significantly larger, which is evident from the corresponding areas of the posterior regression lines. Summary statistics of the posterior parameters are shown in Table \ref{tab:linreg_stats}. \textcolor{black}{Finally, this case is equivalent to a rule-based version with a prior of the form $\operatorname{Beta}(1,1)$ for the $\rules | \pars$, as shown in Figure \ref{fig:Betadist} (right).}

\subsubsection{Rule-based Bayesian linear regression with strict rules (RBLR-s)}

\begin{figure}
  \includegraphics[width=\columnwidth]{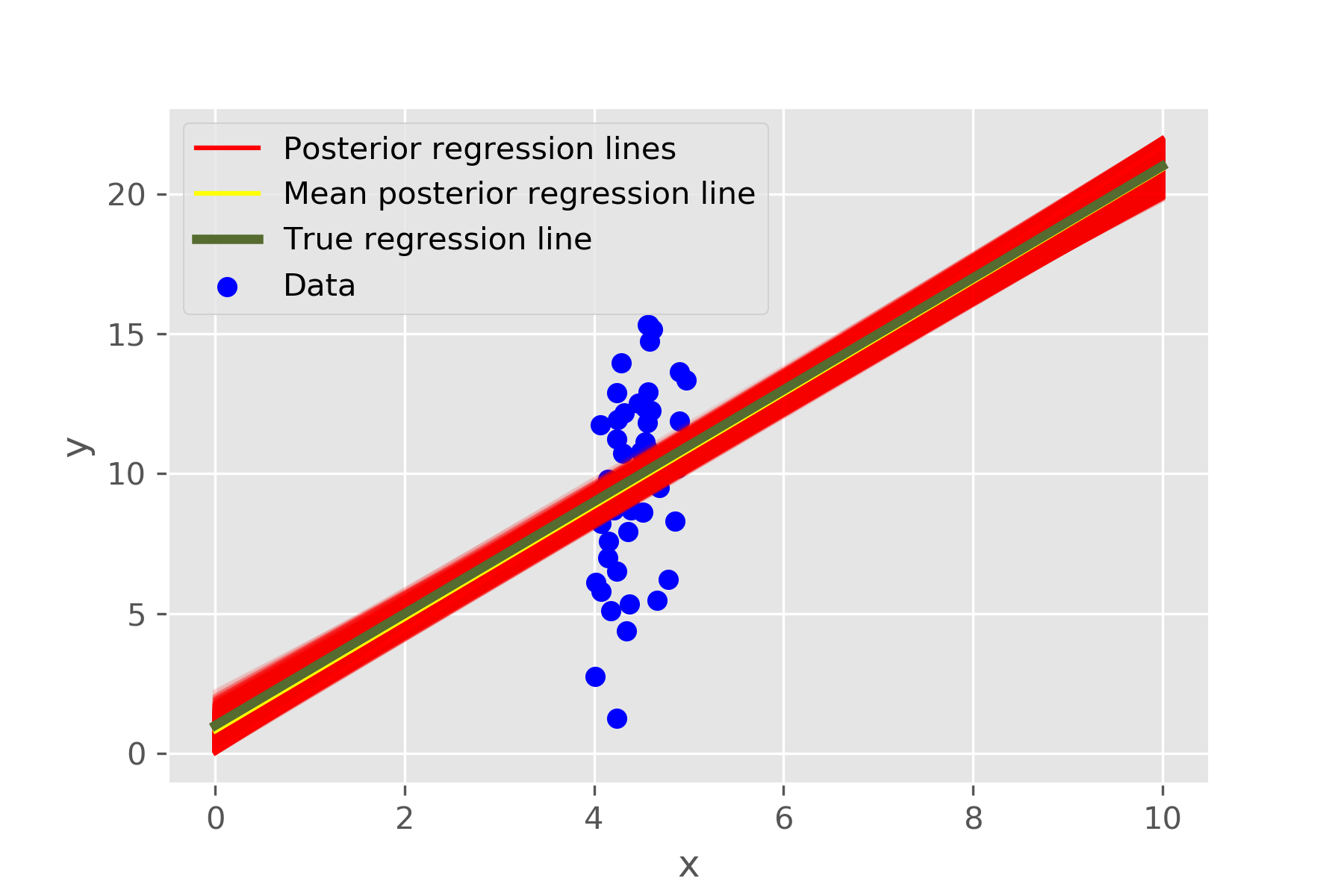}
\caption{Posterior regression lines for the rule-based regression (strict rules). The red lines are realisations of (thinned) samples of the MCMC chain.}
\label{fig:linreg_rulepost}
\end{figure}

We now follow simple intuitions: (i) if the $x$ value is small, then the $y$ value should be small and (ii) if the $x$ value is large, then the $y$ value should be large. We assign specific values that represent the above intuitions in a mathematical manner. The rule base becomes:

\begin{align*}
    R_1&: \text{if} \quad 0 \leq x \leq 1,\quad \text{then} \quad 0 \leq y \leq 4,\\
    R_2&: \text{if} \quad 9 \leq x \leq 10,\quad \text{then} \quad 18 \leq y \leq 22.
\end{align*}
and the composite rule base ($R_\text{comp}$) is given by
\begin{equation*}
    R_\text{comp} := R_1 \land\ R_2 .
\end{equation*}

We also use the distribution $\rules | \pars \sim \operatorname{Beta}(1,100)$ \textcolor{black}{as shown in Figure \ref{fig:Betadist} (left)}, that reflects the level of our confidence in the rules. The form of this distribution\textcolor{black}{, which is a very informative prior,} indicates high confidence in the rule base.


\textcolor{black}{The analytical steps for each MCMC iteration are presented in Algorithm \ref{alg:sampling steps}.}

\begin{algorithm}
\SetAlgoLined
\For{each iteration}{
\textcolor{black}{Sample new values of $\alpha$ and $\beta$\;
Construct discretizations of the inputs. Take $20$ equally-spaced points between $0$ and $1$ (antecedent of the first rule) and $20$ equally-spaced points between $9$ and $10$ (antecedent of the second rule)\;
Compute the outputs for the parameter values of step 1 and the discretizations of step 2\;
Calculate how many of these points violate the corresponding consequents\;
Calculate the ratio of the points that violate the rules over the sum of all ($40$) points\;
Compute $\rules | \pars$ (here $\sim \operatorname{Beta}(1,100)$\;
Calculate the un-normalized posterior as the product of the prior, the likelihood and the quantity $\rules | \pars$ from the previous step}\;
}
 \caption{\textcolor{black}{Analytical sampling steps}}\label{alg:sampling steps}
\end{algorithm}

The results are shown in Figure \ref{fig:linreg_rulepost}. We can observe that, even though there is no apparent difference from the mean prediction when compared to the standard Bayesian regression case (in both cases the mean prediction is very close to the true regression line), the inclusion of the rules reduced the uncertainty significantly, as reflected by the considerably narrower range of the posterior regression lines. This is also evident from the summary statistics in Table \ref{tab:linreg_stats}, where the means of the posterior parameters are very similar to those of the standard method, but the standard deviations are significantly smaller. 

\subsubsection{Rule-based Bayesian linear regression with non-strict rules (RBLR-l)}

\begin{figure}
  \includegraphics[width=\columnwidth]{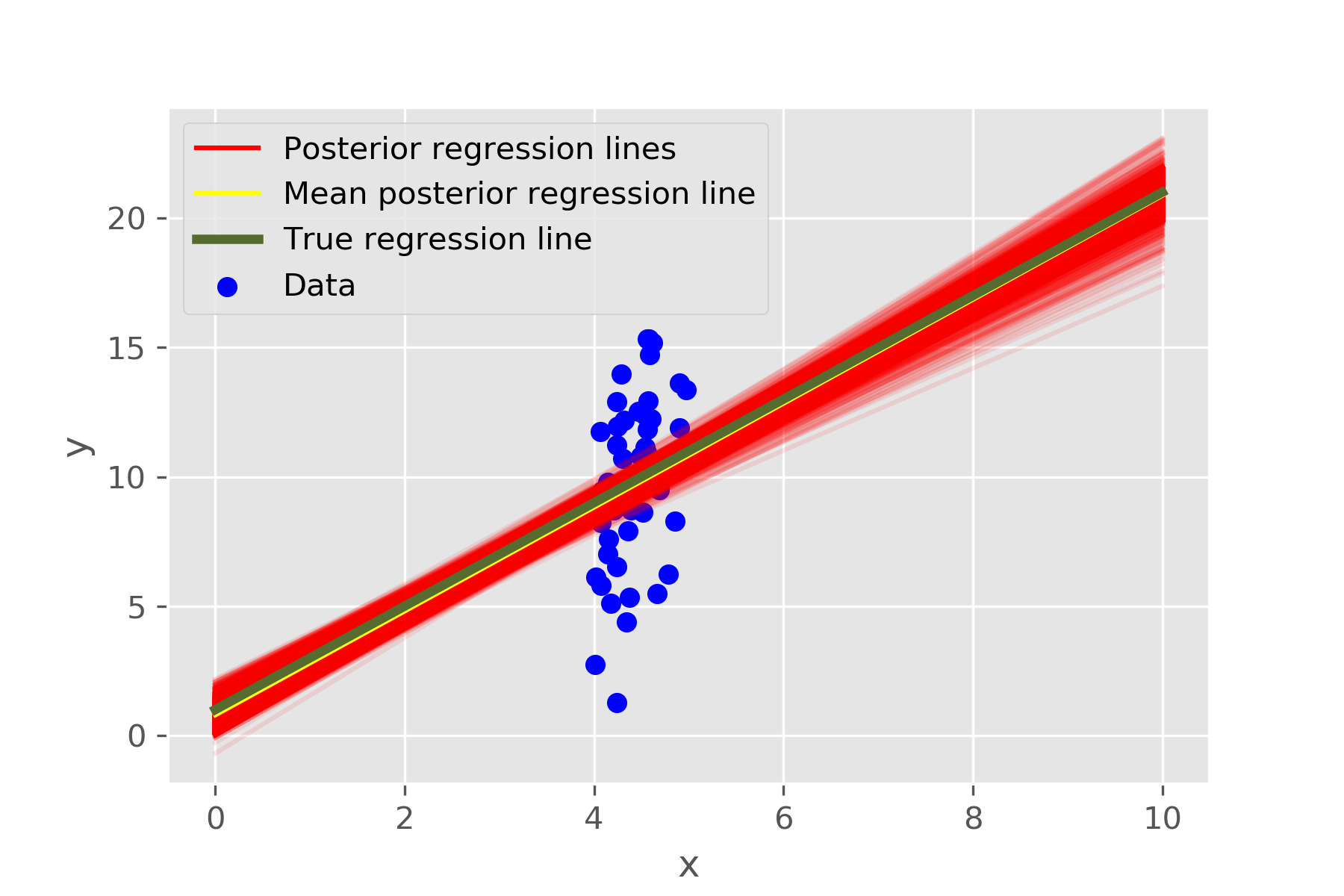}
\caption{Posterior regression lines for the rule-based regression (non-strict rules). The red lines are realisations of (thinned) samples of the MCMC chain. The yellow line is the mean posterior regression line and the green line is the true regression line.}
\label{fig:linreg_rulepost_b5}
\end{figure}

The rule base for this section is the same as the one in the preceding section, however we modify the conditional distribution to $\rules | \pars \sim \operatorname{Beta}(1,5)$ \textcolor{black}{(Figure \ref{fig:Betadist} (middle))}, which reflects lower confidence for the aforementioned rules.

We can see in the Figure \ref{fig:linreg_rulepost_b5} results that the main difference with respect to the two previous cases is, again, the range of the uncertainty level, which is larger than in the case with the strict rules, but narrower than in the case without rules. In practice, the non-strict rules permit the acceptance of more MCMC samples than the stricter rules, by imposing a softer penalty. In contrast, many samples that were accepted with the the standard Bayesian linear regression still get rejected. Even though the difference between the non-strict and strict rules' variations is noticeable in the posterior plots, the summary statistics in Table~\ref{tab:linreg_stats} are very similar, with only slightly larger standard deviation for the parameter $\beta$.

\subsubsection{Rule-based Bayesian linear regression with hyperparameters (RBLR-h)}\label{subs:lin_hyper}
\begin{figure}
  \includegraphics[width=\columnwidth]{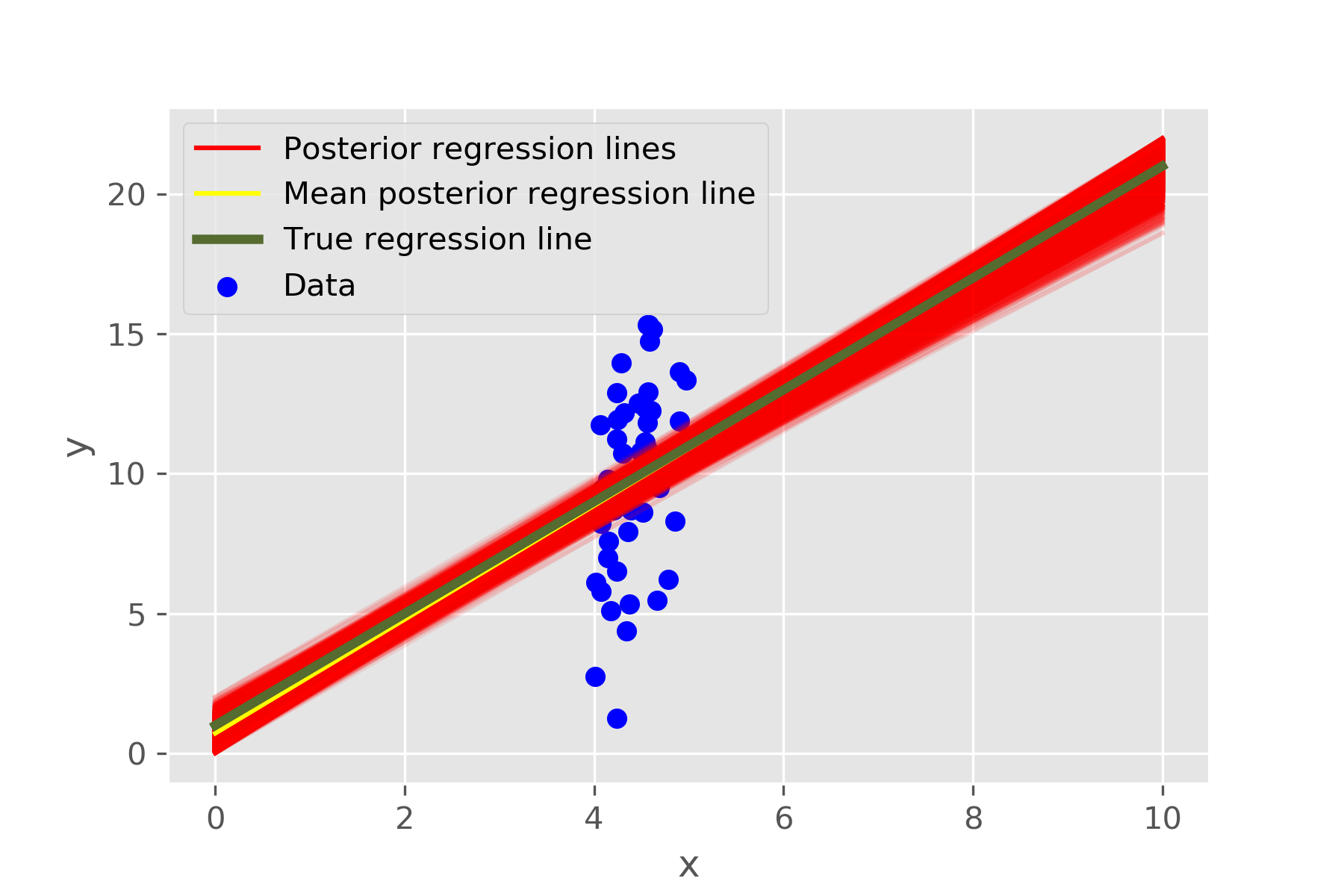}
\caption{Posterior regression lines for the rule-based regression with hyperparameters. The red lines are realisations of (thinned) samples of the MCMC chain. The yellow line is the mean posterior regression line and the green line is the true regression line.}
\label{fig:linreg_hyperulepost}
\end{figure}

We now introduce hyperparameters for the rules themselves. This is useful in a practical setting where expert elicitation results in the form and structure of such rules but precise numeric estimates of the parameter ranges in the antecedent and consequent sections of the rules are difficult to define.

The rule base is modified as follows:
\begin{align*}
    R_1&: \text{if} \quad 0 \leq x \leq x_\text{low},\quad \text{then} \quad 0 \leq y \leq y_\text{low},\\
    R_2&: \text{if} \quad x_\text{high} \leq x \leq 10,\quad \text{then} \quad y_\text{high} \leq y \leq 22,
\end{align*}
where the composite rule base $R_\text{comp}$ is given by
\begin{equation*}
    R_\text{comp} := R_1 \land\ R_2
\end{equation*}
and the hyperparameters are assigned the following prior distributions:
\begin{align*}
    x_\text{low}  &\sim \mathcal{N}(1.5, 0.5^2), \\
    x_\text{high} &\sim \mathcal{N}(8.5, 0.5^2), \\
    y_\text{low}  &\sim \mathcal{N}(4.5, 0.5^2), \\
    y_\text{high} &\sim \mathcal{N}(18.5, 0.5^2).
\end{align*}

\textcolor{black}{It is worth noting that if concrete information about the ranges of $x$ and $y$ are known, truncated versions of the priors above, or other type of distributions (e.g. closed Uniform) might be more favorable.}

The posterior result is shown in Figure \ref{fig:linreg_hyperulepost}. We can observe once more that the main difference with respect to the other cases lies on the uncertainty, which, as with the case with the non-strict rules, is somewhere between the standard Bayesian linear regression case and the case of the rule-based Bayesian linear regression with strict rules.
The introduction of the hyperparameters in the rules acts as a form of regularisation. Both the RBLR-s and RBLR-l are alternative methods for introducing a level of doubt concerning the validity of the rule base.

The results in Table \ref{tab:linreg_stats} indicate that RBLR-h is the only variation with significantly different posterior parameter means, while the standard deviations are in the lower end. This result can be attributed to the effect of the hyperparameters ($x_\text{low}, x_\text{high}, y_\text{low}, y_\text{high}$) that changes the shape of the posterior distribution and redistributes the uncertainty of the system.

\textcolor{black}{It is worth noting that there are significant differences among the true and estimated values of the linear regression parameters. Specifically we can observe noteworthy underestimation of the intercept mean and the parameter variances. The main reason for this is the limited sample that is considered 'observed data', which prohibits us from retrieving the original values. Instead, our goal is to approach the real mean values with relatively low uncertainty. This is achieved in all models, since the true means are within one standard deviation of the estimated values.}

\subsubsection{Remarks}

There are two main takeaways from the regression case study. First, the use of rule-based Bayesian regression can lead to a significant reduction in uncertainty. The introduction of a meaningful rule basis can be a powerful tool that can help introduce expert intuition into a model that otherwise could have only had a post-hoc effect. This additional information can be the cause for a decisive reduction in uncertainty, which can play an important role in decision making. Second, the method is flexible as variations can be employed to articulate expert information and indicate the level of confidence.

\subsection{One-dimensional velocity advection equation}

The velocity advection equation governs transport of momentum by bulk motion. Its form, with a forcing function, in one dimension is
\begin{equation}\label{eq:Burgers1D}
    \frac{\partial u}{\partial t} + u \frac{\partial u}{\partial x} = f(x,t;a,\phi),
\end{equation}
where $u(x,t)$ is the velocity, $x$ denotes the position, $t$ the time, $f(x,t;a,\phi)$ is an external forcing term with amplitude and phase parameters $a$ and $\phi$, respectively.

For the analyses that follow we fit third-degree B-spline models with $10$ knots, while for sampling we use the PyMC3 sequential Monte Carlo (SMC) variation, which is a mixture of the Transitional Markov Chain Monte Carlo (TMCMC) \citep{ching2007transitional} and Cascading Adaptive Transitional Metropolis In Parallel (CATMIP) \citep{minson2013bayesian} algorithms. The number of draws is set to \num{10000}, which in this implementation also corresponds to the number of chains. For the posterior plots we use a thinning of $10$.

\subsubsection{Data}

\begin{figure}
  \includegraphics[width=\columnwidth]{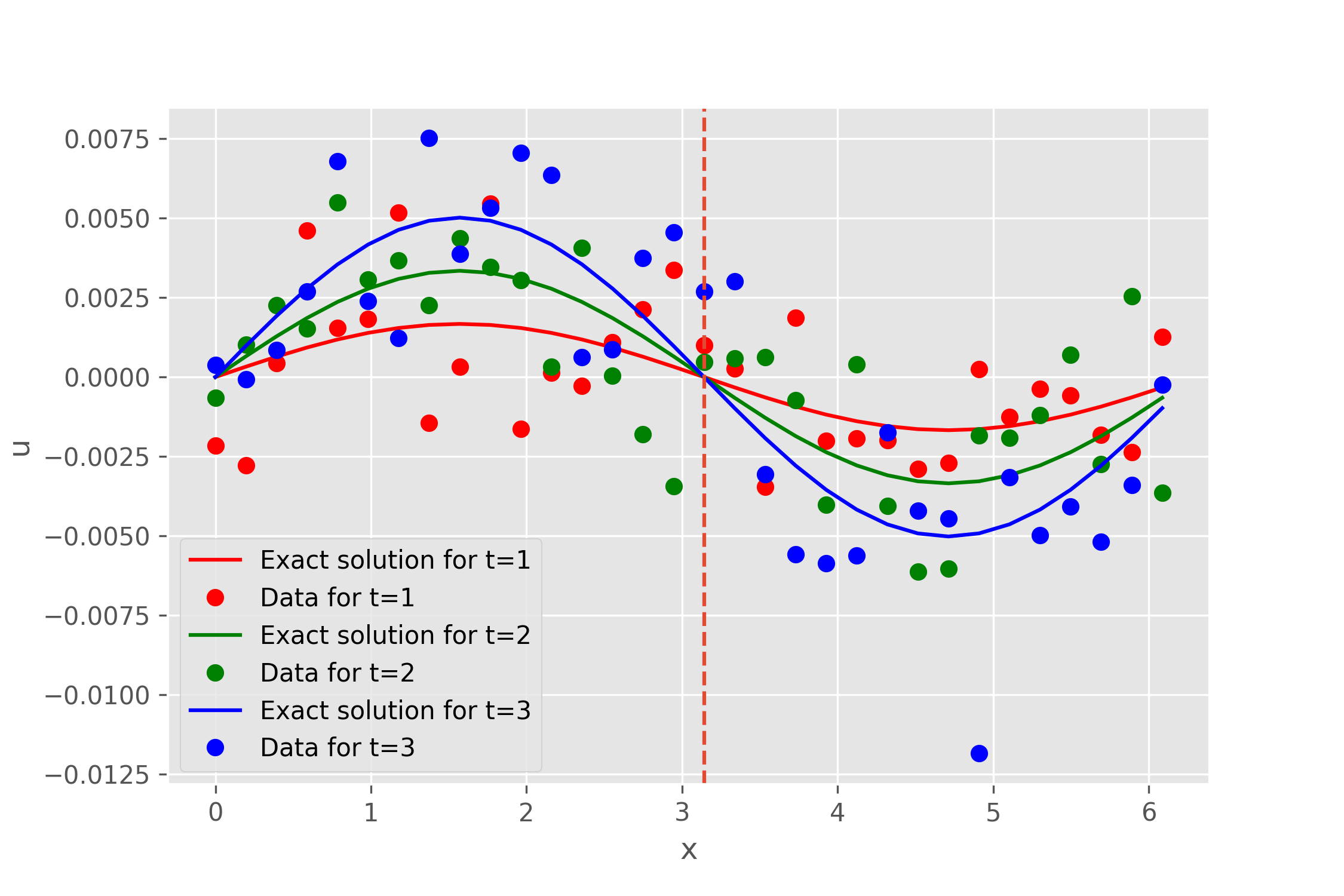}
\caption{One-dimensional advection data for snapshots $t=1$, $2$ and $3$ with corresponding exact solution lines in red, green and blue respectively. The red dashed vertical line denotes the point where all curves intercept ($x = \pi$).}
\label{fig:1DBurgers_data}
\end{figure}

We construct the data from a one-dimensional advection equation with amplitude $a=0.01$ and phase $\phi = \pi$. We extract the data for three different snapshots $t_j$ (corresponding to $t = 1$, $2$, $3$), before adding a Gaussian error with a standard deviation of $0.002$. In mathematical form:
\begin{align*}
    y &= u(x, t) + \epsilon\\
    \epsilon &\sim \mathcal{N}(0,0.002^2).
\end{align*}
The data consist of $96$ points ($32$ values for each snapshot). All the data are shown in a single plot in Figure \ref{fig:1DBurgers_data}.

\subsubsection{Bayesian B-splines regression}

\begin{figure}
  \includegraphics[width=\columnwidth]{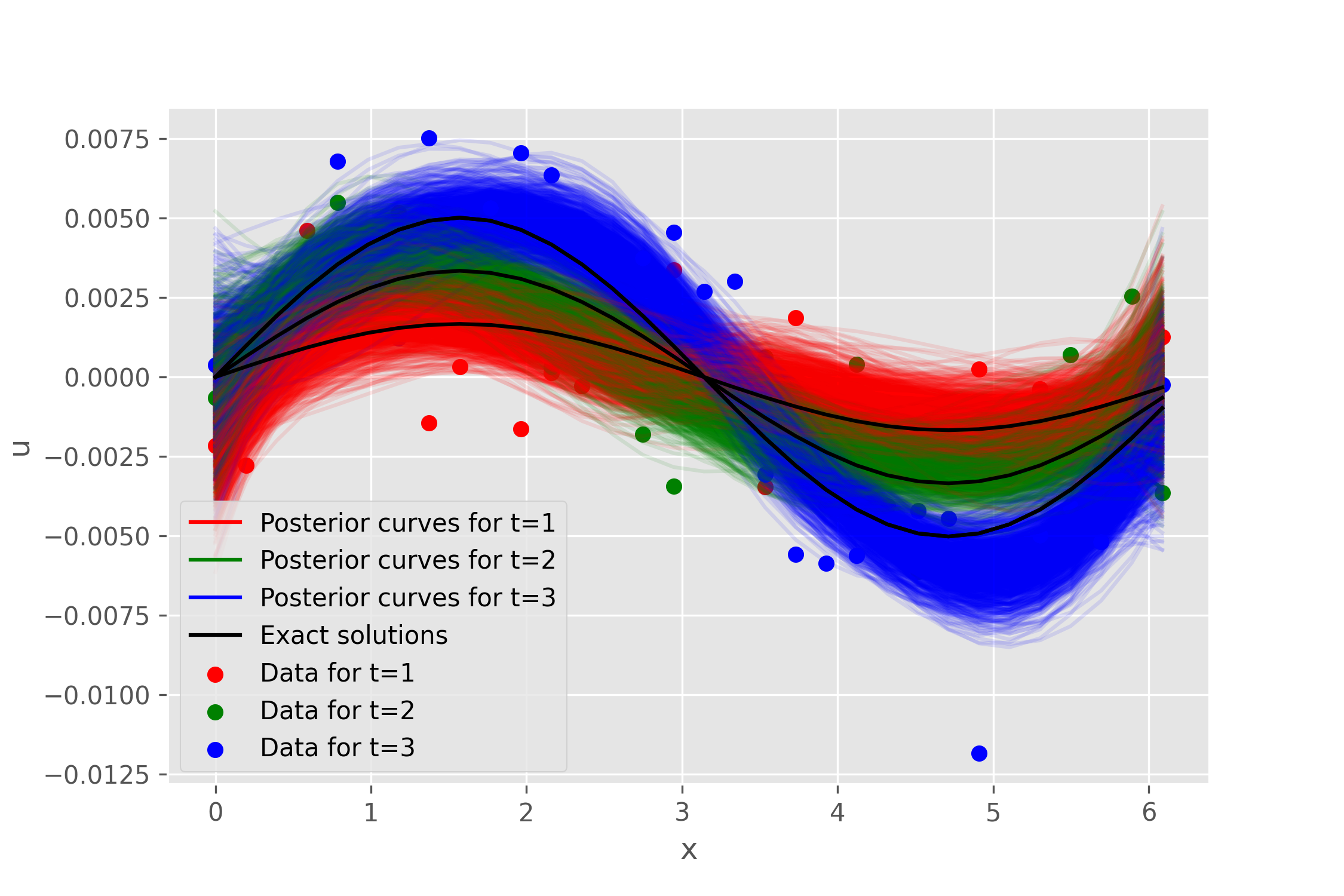}
\caption{Posterior curves for the standard regression. The red, green and blue lines are (thinned) samples of the MCMC chain. The black lines denote the exact solutions.}
\label{fig:1Dburgers_post}
\end{figure}

\begin{figure}
  \includegraphics[width=\columnwidth]{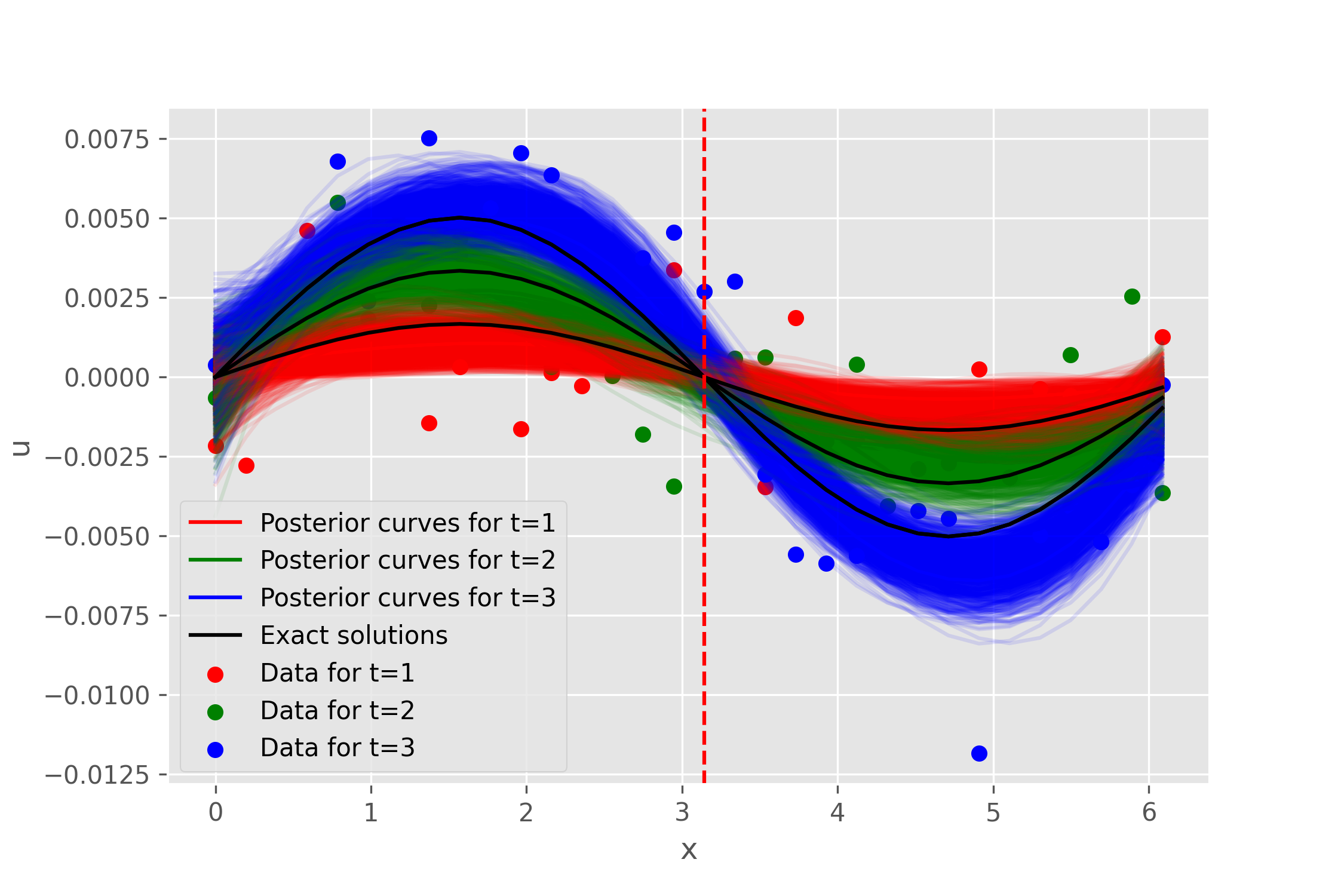}
\caption{Posterior curves for the rule-based regression. The red, green and blue lines are (thinned) samples of the MCMC chain. The black lines denote the exact solutions and the red dashed vertical line denotes the point where the rules change ($x = \pi$).}
\label{fig:1Dburgers_rulepost}
\end{figure}

The parameterisation of the B-splines is \textcolor{black}{derived from \citep{splines_stan, splines_pymc3}, and is known to enforce smoothness and avoid overfitting:}
\begin{equation*}
    a_k = a_0 + \sigma_a \sum_{i=0}^k \Delta a_i,
\end{equation*}
\textcolor{black}{where $k$ is the B-spline degree} and the corresponding priors are:
\begin{align*}
    a_0 &\sim \mathcal{N}(0, 0.1^2), \\
    \Delta a_i &\sim \mathcal{N}(0, 5^2), \\
    \sigma_a &\sim \operatorname{HalfCauchy}(0.1).
\end{align*}

The results of the standard Bayesian B-splines regression are shown in Figure \ref{fig:1Dburgers_post}. We observe that the posterior curves provide predictable results, with the uncertainty ranges of the posterior sets covering the true corresponding curves. There is a very large overlap of the curves that correspond to different snapshots, especially in the center of the plot ($x= \pi$). Finally, in the left and right edges of the posterior plots we observe extreme changes in curvature that do not appear in the corresponding exact solutions.

\subsubsection{Rule-based Bayesian B-splines regression}

From the form of the advection equation, we expect the true curve of each snapshot to be higher than the corresponding curve of the previous snapshot within the interval $(0, \pi)$. Conversely, within the interval $(\pi, 2\pi)$ the true curve of each snapshot should be lower than the curve of the previous snapshot. We also know that for $t=0$ the form of the equation corresponds to the straight line $u=0$. Finally, since the $x$-axis corresponds to a periodic $2\pi$ range, it is expected that the first $x=0$ and last $x=2\pi$ points for each curve should have the same value (in this case $y=0$). We can combine the above intuition into a set of rules:
\begin{align*}
    R_1'&: \text{if} \quad 0 \leq x \leq \pi,\quad \text{then} \quad y_1 \geq 0\\
    R_1&: \text{if} \quad 0 \leq x \leq \pi,\quad \text{then} \quad y_j \leq y_{j+1}, \quad \text{for} \quad j = 1,2\\
    R_2'&: \text{if} \quad \pi \leq x \leq 2\pi,\quad \text{then} \quad y_1 \leq 0\\
    R_2&: \text{if} \quad \pi \leq x \leq 2\pi,\quad \text{then} \quad y_j \geq y_{j+1}, \quad \text{for} \quad j = 1,2\\
    R_3&: \text{if} \quad x_{\text{low}} = 0 \quad \text{AND} \quad x_{\text{high}} = 2\pi,\quad \text{then}\\
    &\quad \|y_{\text{low}} - y_{\text{high}}\| \leq 0.001.
\end{align*}
and the composite rule base $R_\text{comp}$ is given by
\begin{equation*}
    R_\text{comp} := R_1' \land\ R_1 \land\ R_2' \land\ R_2 \land\ R_3 .
\end{equation*}
Note that for numerical reasons we modify the last rule to reflect approximate periodic equality.
For the rule conditional distribution, we use $\rules | \pars \sim \operatorname{Beta}(1,100)$.

As shown in the results of Figure \ref{fig:1Dburgers_rulepost}: (i) the reduction in the uncertainty is visible for all three sets of curves that correspond to the different snapshots, (ii) the overlap in the middle of the $x$-axis (close to $x=\pi$) is reduced to a minimum, and (iii) the extreme curvature behaviour of the plots on the edges has decreased compared with the corresponding standard regression results.

\subsubsection{Remarks}
The first key outcome from this application is that, if required, the antecedents and consequents within a rule base can comprise a more complex form by combining operations (AND, OR, NOT). Additionally, we can supply the rules using intuition from multiple sources. On this occasion, we used the physics as the primary source of expert knowledge.

\subsection{Two-dimensional advection–diffusion equation}

We now consider a two-dimensional advection–diffusion equation that is present in many engineering design problems:
\begin{equation}\label{eq:Burgers2D}
    \frac{\partial c}{\partial t} + u \frac{\partial c}{\partial x} + v \frac{\partial c}{\partial y} = D \left(\frac{\partial^2 c}{\partial x^2} + \frac{\partial^2 c}{\partial y^2} \right),
\end{equation}
where $c$ is a concentration, $x$ and $y$ are spatial coordinates, and $D$ is a diffusion coefficient. We are interested in the concentration profile for different spatio–temporal configurations. Our goal is to construct a GP emulator that acts as a good approximation of full equation.

Since this is a GP model, we will use an optimisation technique in order to find the maximum a posteriori (MAP) estimate of the penalised maximum likelihood, Equation~\ref{eq:mod_marg_like}. We will then use the mean of the predictive posterior distribution in order to reconstruct each of the snapshots and compare them with their true counterparts. Specifically, we use the Nelder–Mead \citep{nelder1965simplex} optimisation algorithm.

For the GP, we use a zero mean function and the automatic relevance determination (ARD) variation of the Matérn~3/2 kernel \citep{bishop2006pattern}, which incorporates a separate parameter for each input variable, adding to the flexibility of the kernel:
\begin{equation}
    k(\mathbf{x}, \mathbf{x'}) = 
    \zeta^2 \left(
        1 + \frac{\sqrt{3(\mathbf{x}-\mathbf{x}')^2}}{\mathbf{l}}
    \right)
    \exp\left(-\frac{\sqrt{3(\mathbf{x}-\mathbf{x}')^2}}{\mathbf{l}}\right),
\end{equation}
where $\mathbf{l}$ is a three-dimensional vector\textcolor{black}{, accounting for the two spatial and one temporal components. Note that the kernel is not separable}. We also take into account the variance of the system $\sigma$: for priors we use
\begin{align*}
    \mathbf{l} &\sim \operatorname{Gamma}(1, 1) \\
    \zeta &\sim \operatorname{HalfCauchy}(1) \\
    \sigma &\sim \operatorname{HalfCauchy}(1),
\end{align*}
where $\operatorname{HalfCauchy}$ is the truncated Cauchy distribution in which only values to the right of the peak have nonzero density.

\subsubsection{Data}

\begin{figure*}
    \centering
    \mbox{
    \subfigure[True frames.]{\includegraphics[width=\textwidth]{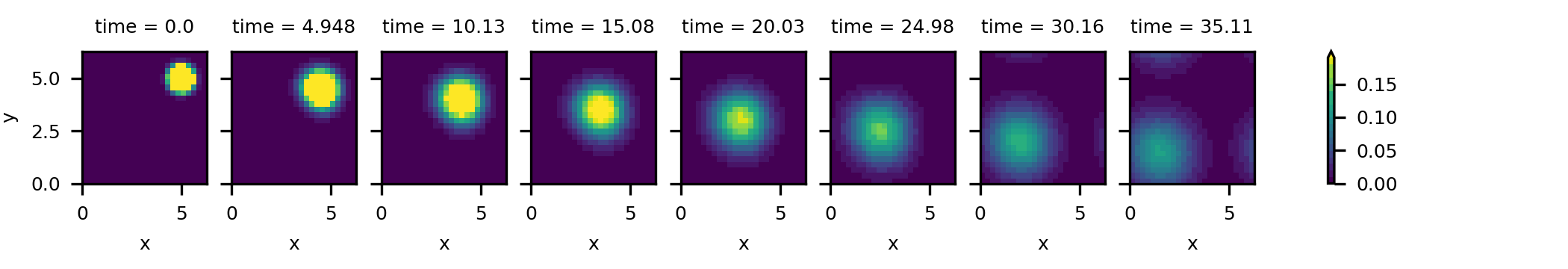}\label{fig:2Dburg_post_true}
    }}\\
    \mbox{
    \subfigure[Frames reconstructed with the GP regression model.]{\includegraphics[width=\textwidth]{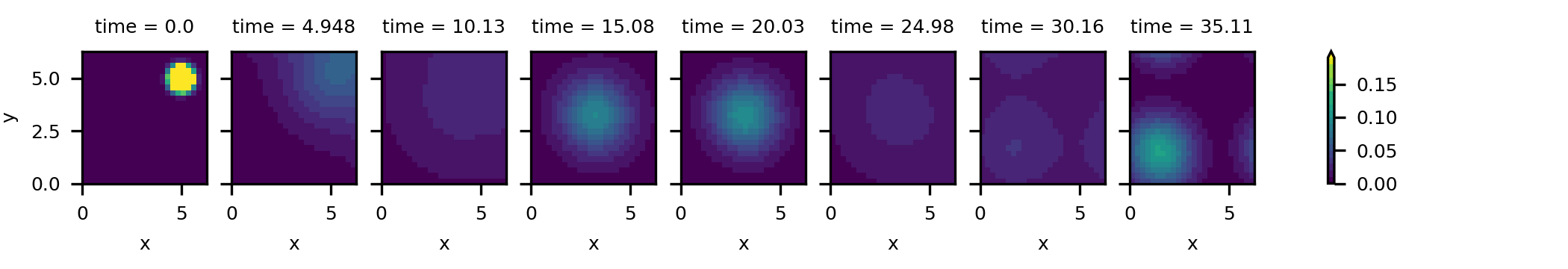}\label{fig:2Dburgers_post_norule}}
    }\\
    \mbox{
    \subfigure[Frames reconstructed with the rule-based GP regression model.]{\includegraphics[width=\textwidth]{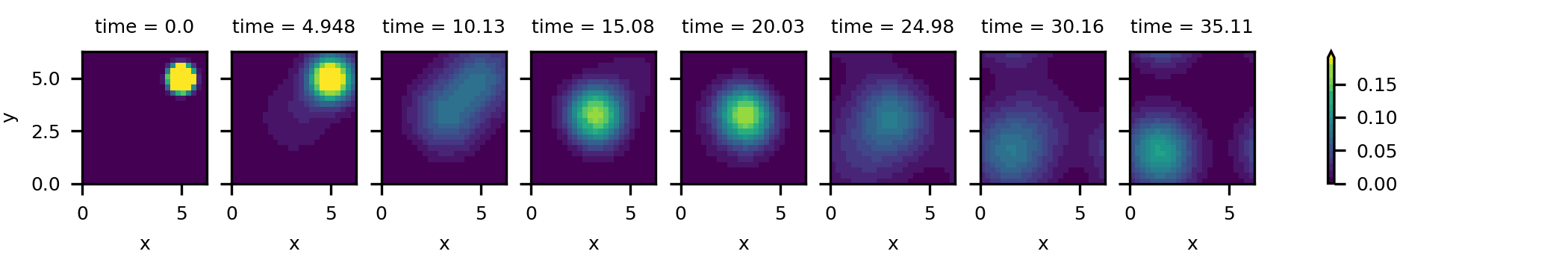}\label{fig:2Dburgers_post_rule}}
    }\\
    \caption{Indicative concentration fields for selected time steps. The first ($t = 0$) and the last ($t=35.11$) frames correspond to input snapshots. The third input snapshot ($t = 17.67$) is between the fourth and fifth frame.}
    \label{fig:2Dburgers_allres}
\end{figure*}

\begin{figure}
  \includegraphics[width=\columnwidth]{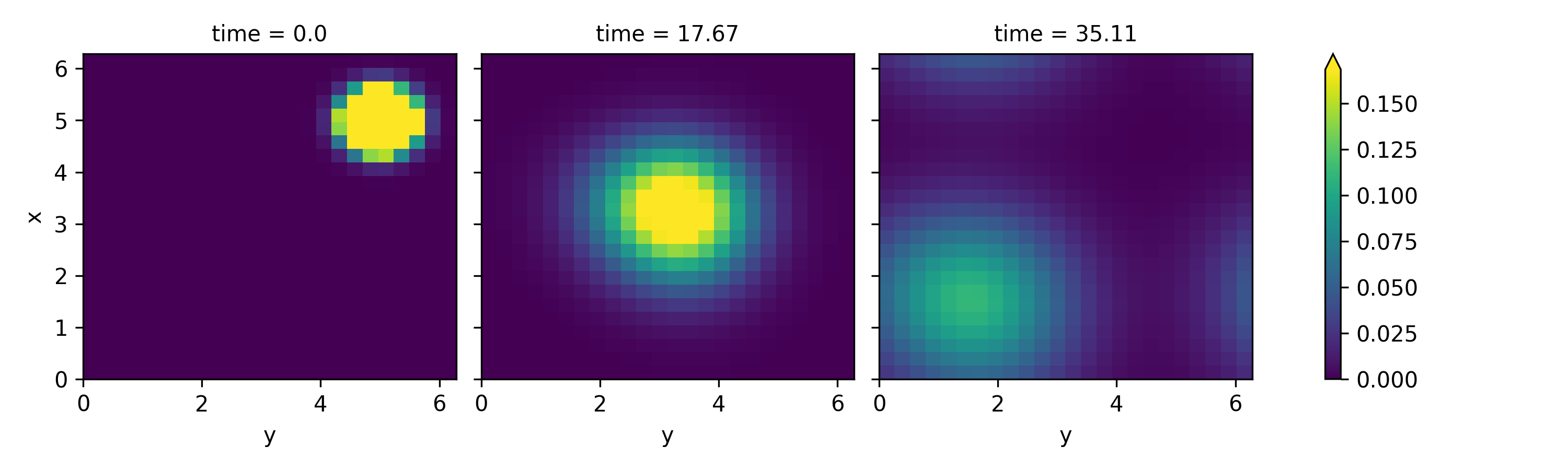}
\caption{The three frames (at times 0,17.47, 35.11) that are used as input for the GP, for the reconstruction of the two-dimensional advection–diffusion solution.}
\label{fig:2Dburgers_input_frames}
\end{figure}

We use a $24 \times 24$ discretisation grid of the two-dimensional interval $[0, 2\pi]\times[0, 2\pi]$. The starting concentration is constructed as a bivariate Gaussian distribution with mean $(5,5)$ and covariance $0.1 I_2$, i.e., $\mathcal{N}(\mu = (5,5), \Sigma = 0.1 I_2)$. The initial velocity field is set to be constant with the value $-0.1$ in both directions ($(u,v) = (-0.1,-0.1)$). In practice, this corresponds to transport of the concentration from the top right corner of the grid to the bottom left. The viscosity value is set equal to $0.02$. Finally, we use $150$ snapshots with a time step of $\SI{\sim 0.235}{s}$ (the total duration spans from $t=0$ to $\SI{\sim 35.11}{s}$).

Indicative solution frames are shown in Figure \ref{fig:2Dburg_post_true}. We note how, as the time increases, the concentration moves down- and left-wards (due to velocity advection), but also increases in radius (due to diffusion).

We now use the limited information from the concentration fields of three snapshots ($t = 0$, $17.67$, $35.11$ shown in Figure \ref{fig:2Dburgers_input_frames} as input for the GPs. The total number of data points is $24\times24\times3 = \num{1728}$, which indicates that, even with a limited number of input snapshots, the number of data points to be processed is large. Further increasing the number of input snapshots adds a heavy computational burden, which in more realistic scenarios could lead to major issues.

\subsubsection{Gaussian process regression}\label{sec:GPR}


Figure~\ref{fig:2Dburgers_allres} shows the true frames (Figure~\ref{fig:2Dburg_post_true}) together with reconstructed frames from the standard GP regression (Figure~\ref{fig:2Dburgers_post_norule}). The frames relatively close to the three input snapshots (the ones that correspond to $t = 0$, $15.08$, $20.03$, $35.11$ in the aforementioned figure) have a reasonably similar concentration profile to that of the corresponding true values. Conversely, for the frames further from the input times ($t=10.13$, $24.98$, $30.16$), the predicted concentration seems almost uniform, and the original profile can no longer be detected.

\subsubsection{Rule-based Gaussian process regression}

\begin{figure}
  \includegraphics[width=\columnwidth]{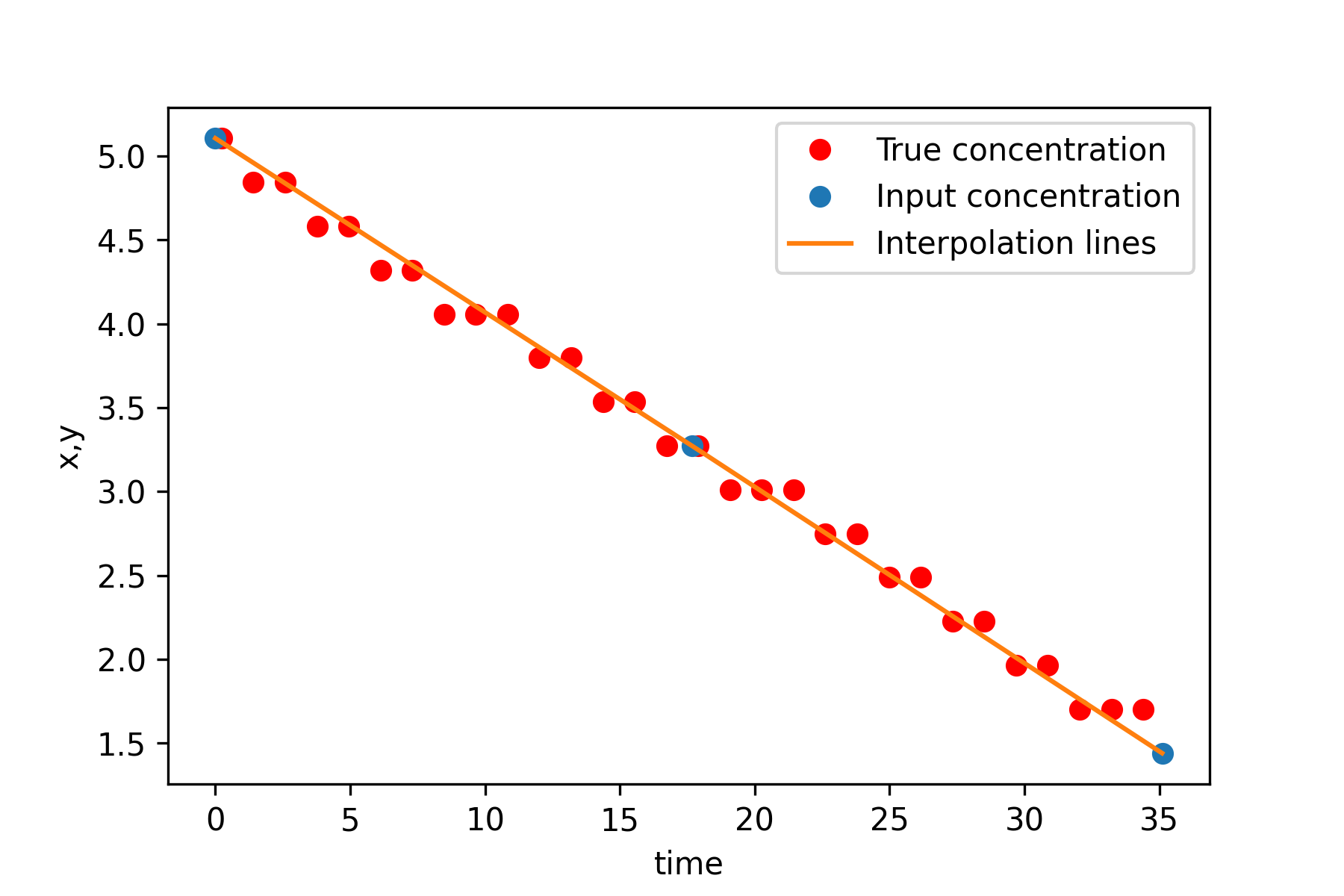}
\caption{Linear interpolation for the rule regarding the position of the point with the highest concentration with respect to time. The $3$ blue points represent the input snapshots, the yellow piece-wise linear curve the interpolation from the inputs and the red points the true values of the highest concentration point position for each of the rule base concentration profiles.}
\label{fig:2Dburgers_rule}
\end{figure}

We construct a rule based on the intuition that the center of the blob (which corresponds to the point with highest concentration) moves from the top right corner to the bottom left corner linearly with respect to time. First, we find the blob centers using the three input snapshots. We then construct the corresponding piece-wise linear interpolation, as shown in Figure~\ref{fig:2Dburgers_rule}. Next, we use the interpolation in order to approximate the blob centers of a pre-specified set of snapshots. The red dots in the figure show the true position of the blob centers, whereas our prediction is the corresponding value that lies on the yellow line. Finally, we set a lower bound for the concentration of the aforementioned blob centers. In our analysis, we use $30$ equally spaced snapshots for the rule base (every $\SI{1.178}{s}$) and we set the value $1$ as the lower bound, which we base on the maximum concentration values of the input frames. The applied rule is hence
\begin{align*}
    R_1: \quad & \text{if} \quad t = 0,1.178,2.356,\dots,\quad \text{then}\\
    &\quad \max(c_t) \geq 1,
\end{align*}
where $c_t$ denotes the concentration at time $t$.
We use the mean as the summary statistic of the distribution $\vy_r | \vy,\vf, \vx$. Finally, for the conditional distribution associated with the rules we use $\rules | \vy_r \sim \operatorname{Beta}(1,100)$.

Results for the same $8$ frames as before are shown in Figure \ref{fig:2Dburgers_post_rule}. Though there is still deviation from the true values, there is considerable progress when we use the rule-based GP regression. The frames that correspond to $t = 10.13,24.98, 30.16$ have a clear concentration profile structure which approximates, at different degrees of accuracy, the true profiles. In Table \ref{tab:gp_metrics} we present the mean squared error (MSE) and mean absolute error (MAE) for the two different approaches for the $3$ input (train column) and the remaining $147$ (test column) original snapshots. The fitting is slightly better with the standard GP regression during training, which likely occurs because the inclusion of the rules adds a penalty to the value that the standard method considers of best fit. During testing, the rule-based GP regression performs better than the standard algorithm, which confirms the intuition from Figure~\ref{fig:2Dburgers_allres}.

\begin{figure}
  \includegraphics[width=\columnwidth]{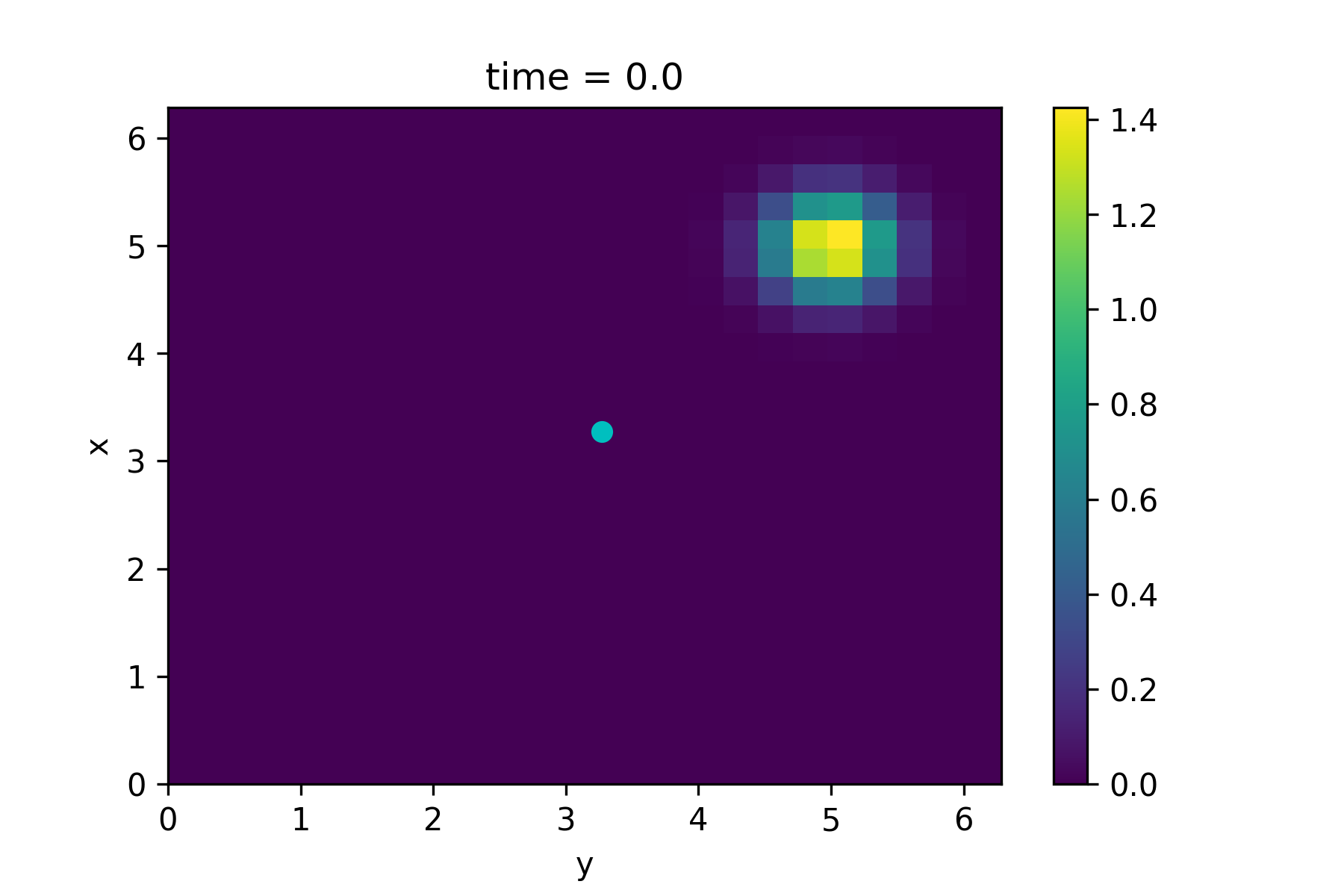}
\caption{The light blue dot in the middle of the frame denotes the point whose time series we focus on.}
\label{fig:2Dburgers_marginal_frame}
\end{figure}

\begin{table*}
    \caption{Mean squared error (MSE) and mean absolute error (MAE) for training ($3$ input snapshots) and testing (remaining $147$ snapshots) for the GP regression (GPR) and the rule-based Gaussian-process regression (RGPR).}
    \centering
    {\begin{tabular}{ccccc}
            \toprule
                   & \multicolumn{2}{c}{Train} & \multicolumn{2}{c}{Test}                                 \\
            \cmidrule(lr){2-3} \cmidrule(lr){4-5}
            Metric & MSE                       & MAE                      & MSE           & MAE           \\ \midrule
            GPR    & \num{2.8657e-9}           & \num{6.2715e-6}          & \num{0.00198} & \num{0.01997} \\
            RGPR   & \num{3.3756e-9}           & \num{6.8051e-6}          & \num{0.00106} & \num{0.01377} \\ \bottomrule
        \end{tabular}}
    \label{tab:gp_metrics}
\end{table*}

\subsubsection{Results for a single grid point}

\begin{figure}
  \includegraphics[width=\columnwidth]{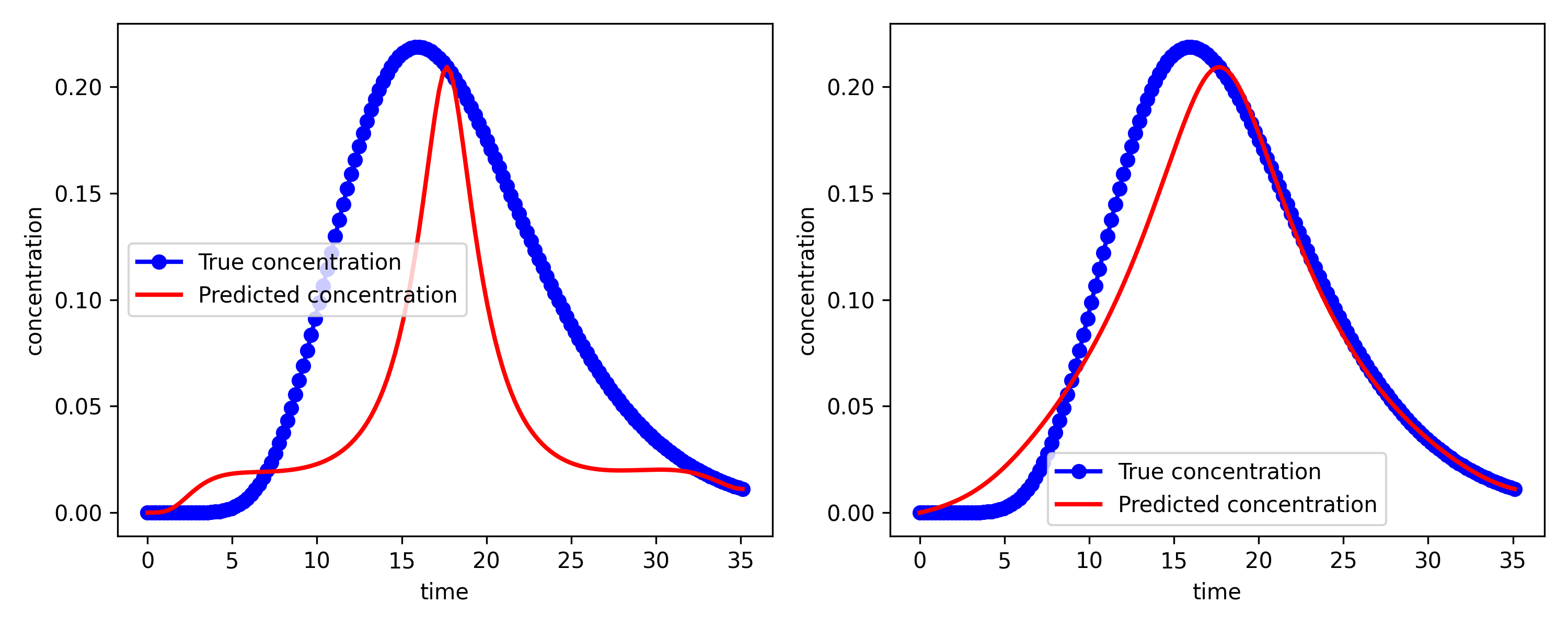}
  \begin{minipage}{.5\textwidth}\centering \subfigure[]{\label{fig:MAP_without_rules}}\end{minipage}
  \begin{minipage}{.5\textwidth}\centering \subfigure[]{\label{fig:MAP_with_rules}}\end{minipage}
\caption{The blue dots denote the real concentration values of the light blue point from Figure~\ref{fig:2Dburgers_marginal_frame} for the entire time duration. The red curve is the corresponding GP MAP prediction (a) without the rules and (b) with the rules.}
\label{fig:2Dburgers_marginal_bothrule_mean}
\end{figure}

Examining the outputs of each regression method for a single grid point allows us to investigate the uncertainty output of the GPs and gain a better understanding of the differences between them. We choose a point in the middle, as shown in Figure \ref{fig:2Dburgers_marginal_frame}.

As discussed in Section \ref{sec:GPR}, the result from the standard GP regression without rules produces accurate predictions only for times close to the input snapshots. This is particularly evident in Figure~\ref{fig:MAP_without_rules}, where the prediction (red curve) coincides with the true values (blue markers) only for the input points, whereas for the rest of the time series, it underestimates the true concentration value. Conversely, with the use of the rule-based GP regression method, the prediction improves significantly even where there is no information from the inputs, as shown in Figure~\ref{fig:MAP_with_rules}.

\begin{figure}
  \includegraphics[width=\columnwidth]{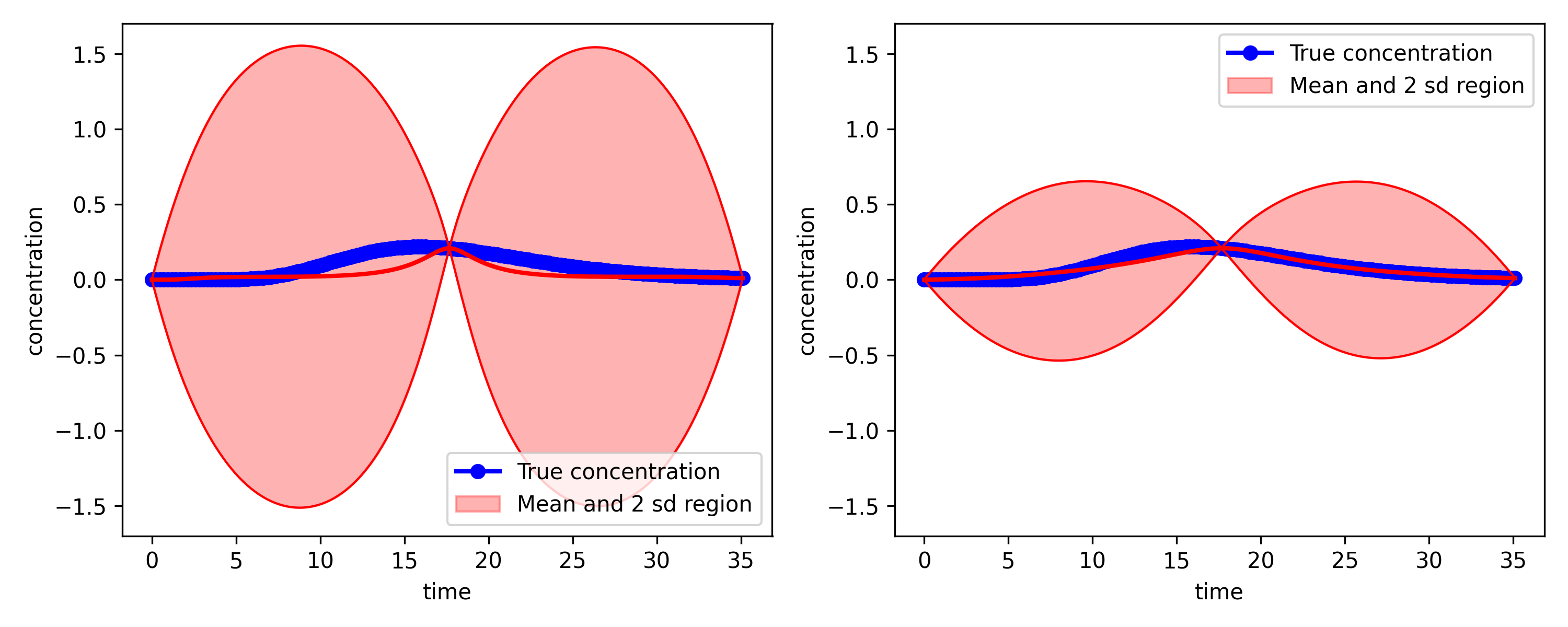}
  \begin{minipage}{.5\textwidth}\centering \subfigure[]{\label{fig:uncertainty_without_rules}}\end{minipage}
  \begin{minipage}{.5\textwidth}\centering \subfigure[]{\label{fig:uncertainty_with_rules}}\end{minipage}
\caption{Same as Figure~\ref{fig:2Dburgers_marginal_bothrule_mean} with the addition of $\pm 2$ standard deviations from the MAP denoted by the shaded area for the case (a) without the rules and (b) with the rules.}
\label{fig:2Dburgers_marginal_bothrule_sd}
\end{figure}

The advantage of examining the time series of a single point is that we can also visualize the uncertainty. In Figure~\ref{fig:uncertainty_without_rules} we plot the prediction with $\pm 2$ standard deviations from the MAP estimate. The uncertainty levels are very high for most of the time series, with the exception of the three input snapshots ($t = 0$, $17.67$, $35.1$) where the uncertainty is very narrow. It is also worth noting that the true concentration values are included within the uncertainty region. The inclusion of the rules, Figure~\ref{fig:uncertainty_with_rules}, decreases the uncertainty range by approximately two thirds and the true concentration is still included within the region. As mentioned in Section \ref{sec:lin_reg}, this reduction in the uncertainty can be critical for the decision making process.

\subsubsection{Remarks}

In this application, we used a rule base that included summary statistics, further demonstrating that the rule component of the methodology is flexible and can potentially accommodate the inclusion of complex expert intuition, that otherwise would be difficult to add to a model. Furthermore, we showed again that the main contribution of the methodology is related to the uncertainty quantification, though there are potential advantages with improvements to point predictions as well.

Regarding the advection–diffusion application, and according to our experiments, the GP, together with the use of various popular kernels, can easily capture the effect of diffusion, but struggles with the movement of the concentration blob, regardless of whether we are using the rule-based version or not. Specifically, as is evident from Figure~\ref{fig:2Dburgers_post_rule}, instead of re-positioning the blob, it diffuses the blob from the previous available input and it reverses the process to arrive at the next input. Nevertheless, our analysis aims to show that the methodology of the rule-based Bayesian regression can improve the result significantly, even in cases where the regression model is not perfectly suitable for the problem that needs to be tackled.

\textcolor{black}{\subsection{Full load electrical power output of a combined cycle power plant}}\label{subs:CCPP}
\textcolor{black}{For the fourth application, our goal is to predict the electrical output of a combined cycle power plant (gas and steam turbines). The data are provided and examined by fitting various Machine Learning models in \citep{Hoyer2020data}.}

\textcolor{black}{For our analysis we fit a multivariate linear regression model. Once again we use the PyMC3 package for sampling, and specifically a single Metropolis - Hastings chain with \num{100000} draws, in addition to a burn-in of \num{30000} iterations and thinning of $100$.}

\textcolor{black}{\subsubsection{Data}}\label{sec:CCPP_data}

\begin{figure}
  \includegraphics[width=\columnwidth]{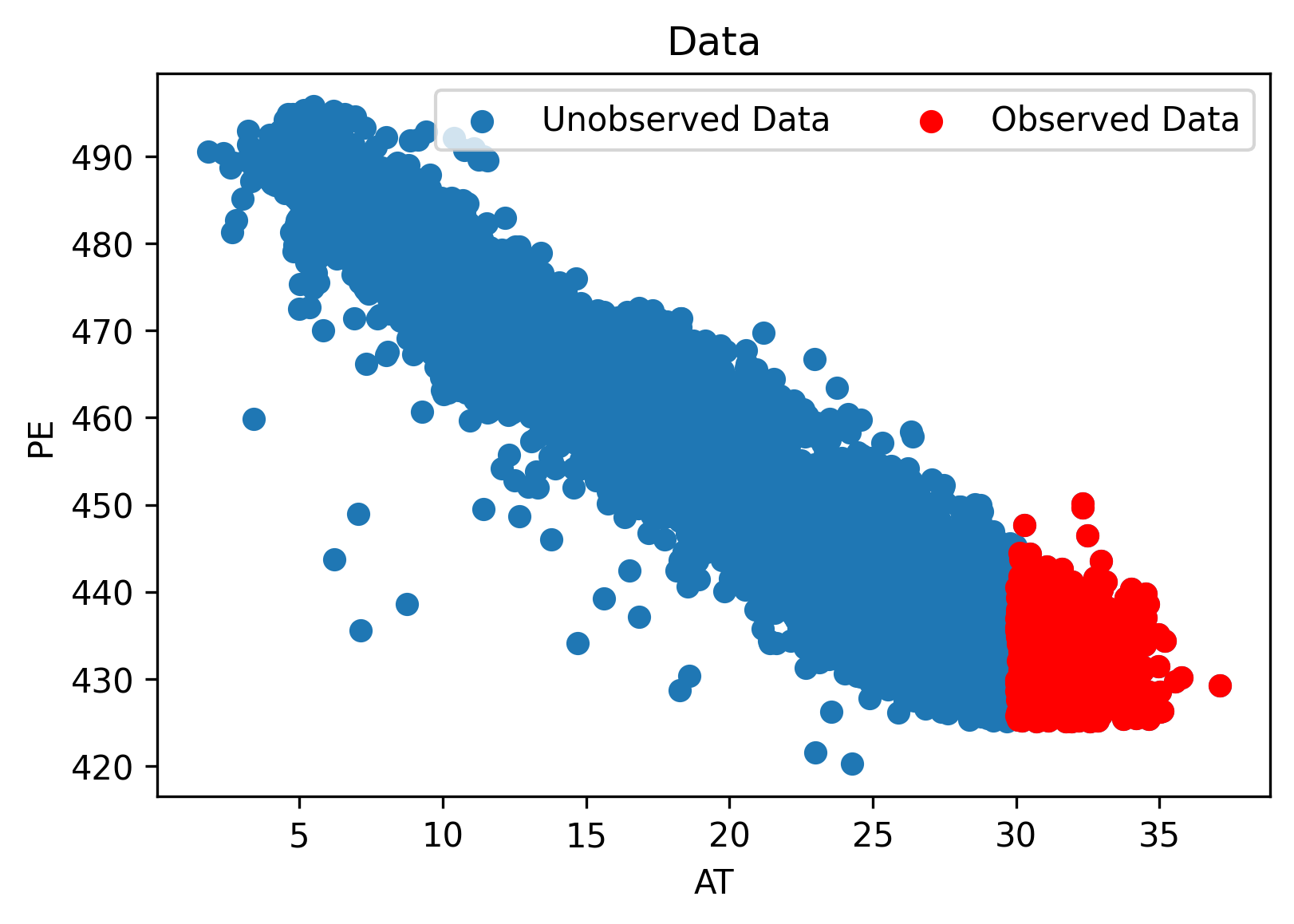}
\caption{\textcolor{black}{Scatterplot of ambient temperature ($AT$) and electrical energy output ($PE$). The red points denote the observed data (which are used for the model fitting) and the blue points the unobserved data (which are used for the model evaluation).}}
\label{fig:CCPP_data}
\end{figure}

\textcolor{black}{The dataset consists of four features; the ambient temperature $AT$, the ambient pressure $AP$, the humidity $RH$ and the vacuum $V$, whereas the electrical energy $PE$ is the single output. We use the part of the dataset where $AT \geq 30^oC$ as the observed values as shown in Figure \ref{fig:CCPP_data}. Effectively, this is equivalent to collection data only during high temperature days (e.g. say only summer season). This leaves $734$ data points from the original $9568$ for the training set. We evaluate the model in the remaining data.}

\textcolor{black}{\subsubsection{Multivariate linear regression}}

\textcolor{black}{We construct a multivariate linear regression model with parameters the coefficients of the features mentioned in the previous sections. This yields:}
\begin{multline*}
    \textcolor{black}{PE = AT_{co}*AT + AP_{co}*AP + RH_{co}*RH +}\\
    \textcolor{black}{V_{co}*V + b + \epsilon,}
\end{multline*}
\textcolor{black}{where $PE$, $AT$, $AH$, $AFDP$ and $GTEP$ are as explained in Section \ref{sec:CCPP_data}, $AT_{co}$, $AP_{co}$, $RH_{co}$ and $V_{co}$ are the corresponding coefficients, $b$ is the intercept and $\epsilon$ Gaussian error with:}
\begin{equation*}
    \textcolor{black}{\epsilon \sim \mathcal{N}(0,1^2).}
\end{equation*}

\textcolor{black}{We use wide Gaussian distributions for the priors:}
\begin{align*}
    \textcolor{black}{AT_{co}, AP_{co}, RH_{co}, V_{co}}  &\textcolor{black}{\sim \mathcal{N}(0, 10^2),} \\
    \textcolor{black}{b} &\textcolor{black}{\sim \mathcal{N}(0, 16^2).}
\end{align*}

\begin{figure}
  \includegraphics[width=\columnwidth]{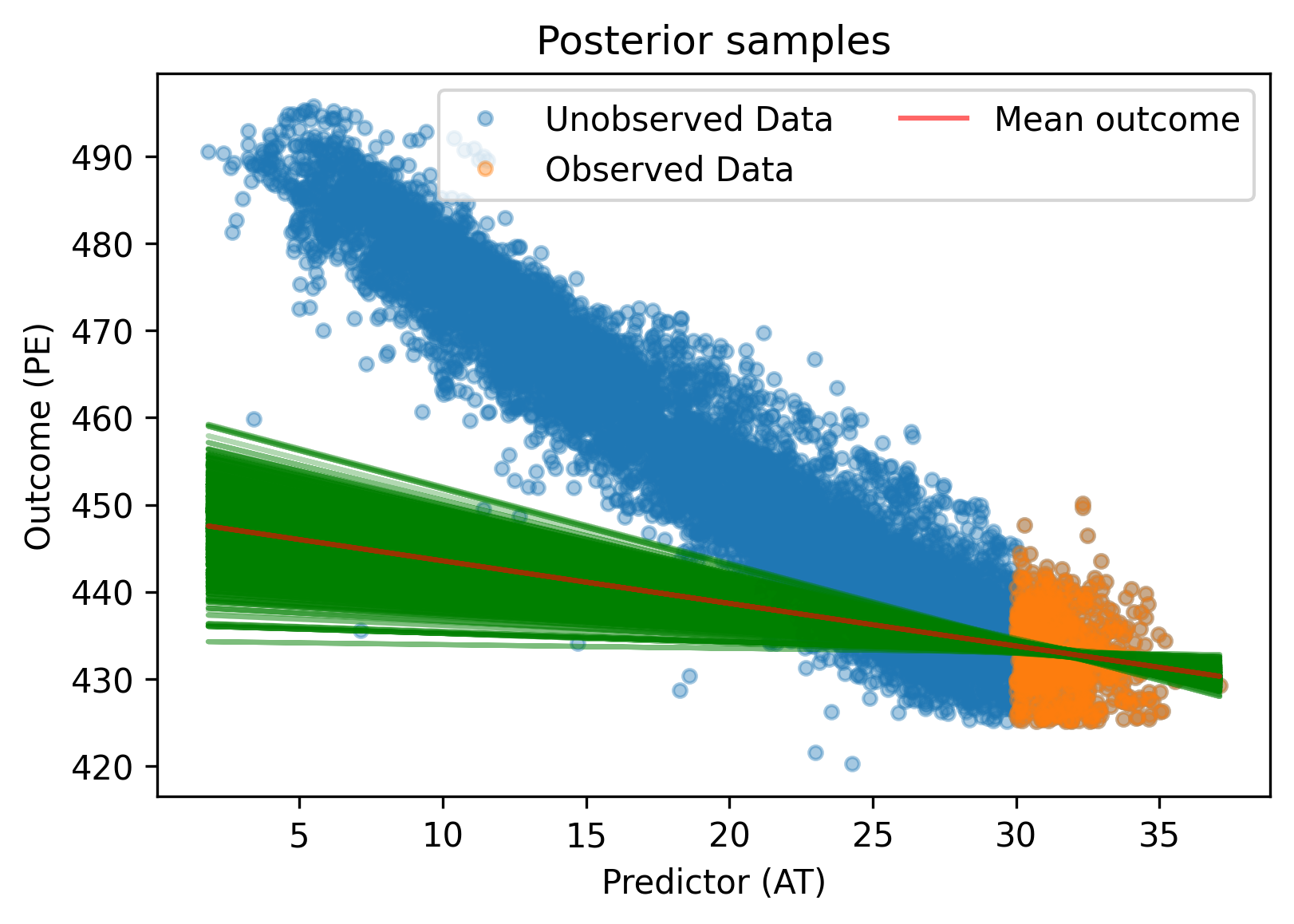}
\caption{\textcolor{black}{Scatterplot of $AT$ and $PE$. The yellow points denote the observed and the blue points the unobserved data. The posterior samples of the linear model without the rules are in green and the mean linear model in dark red.}}
\label{fig:CCPP_norules}
\end{figure}

\textcolor{black}{In Figure \ref{fig:CCPP_norules}, we show the scatterplot of $AT$ and $PE$, along with the posterior samples. The posterior uncertainty increases away from the area of the training data, and the slope is significantly different from the one implied by the data (which has a very strong negative correlation). Evaluation metrics are included in Table \ref{CCPP_metrics}. }

\begin{table}[h]
\caption{\textcolor{black}{Evaluation metrics for the different models.}}
\centering
\begin{tabular}{|c|c|c|}
\hline

\textcolor{black}{Metric/Model} & \textcolor{black}{Without rules}   & \textcolor{black}{With rules}\\ \hline
        \textcolor{black}{RMSE}              & \textcolor{black}{\num{4.051}}     & \textcolor{black}{\num{1.655}}\\ \hline
        \textcolor{black}{MSE}              & \textcolor{black}{\num{16.411}}     & \textcolor{black}{\num{2.740}}\\ \hline
        \textcolor{black}{MAE}              & \textcolor{black}{\num{3.168}}     & \textcolor{black}{\num{1.337}}\\ \hline
\end{tabular}
\label{CCPP_metrics}
\end{table}

\textcolor{black}{\subsubsection{Rule-based Multivariate linear regression}}
\textcolor{black}{In order to derive a rule base for this model, we used expert knowledge from the literature \citep{Hoyer2020data, gonzalez2017effect}. Specifically, we refer to the following statement from the former: \textit{Effect of Ambient Temperature ($AT$): The effect of $AT$ on the performance is the most widely studied subject about gas turbines. This can be appreciated since $AT$ is the most inﬂuential factor showing a correlation around $-0.95$ with $PE$}.}

\textcolor{black}{We can express this in mathematical form as:}
\begin{align*}
    \textcolor{black}{R_1}&\textcolor{black}{: \text{if} \quad AT \leq AT_{low},\quad \text{then} \quad PE \geq PE_{hi}}\\
    \textcolor{black}{R_2}&\textcolor{black}{: \text{if} \quad AT \geq AT_{hi},\quad \text{then} \quad PE \leq PE_{low},}
\end{align*}
\textcolor{black}{where $AT_{low}$ is the $25$th percentile, $AT_{hi}$ is the $75$th percentile, $PE_{low}$ is the $25$th percentile and $PE_{hi}$ is the $50$th percentile of the observed data.}

\begin{figure}
  \includegraphics[width=\columnwidth]{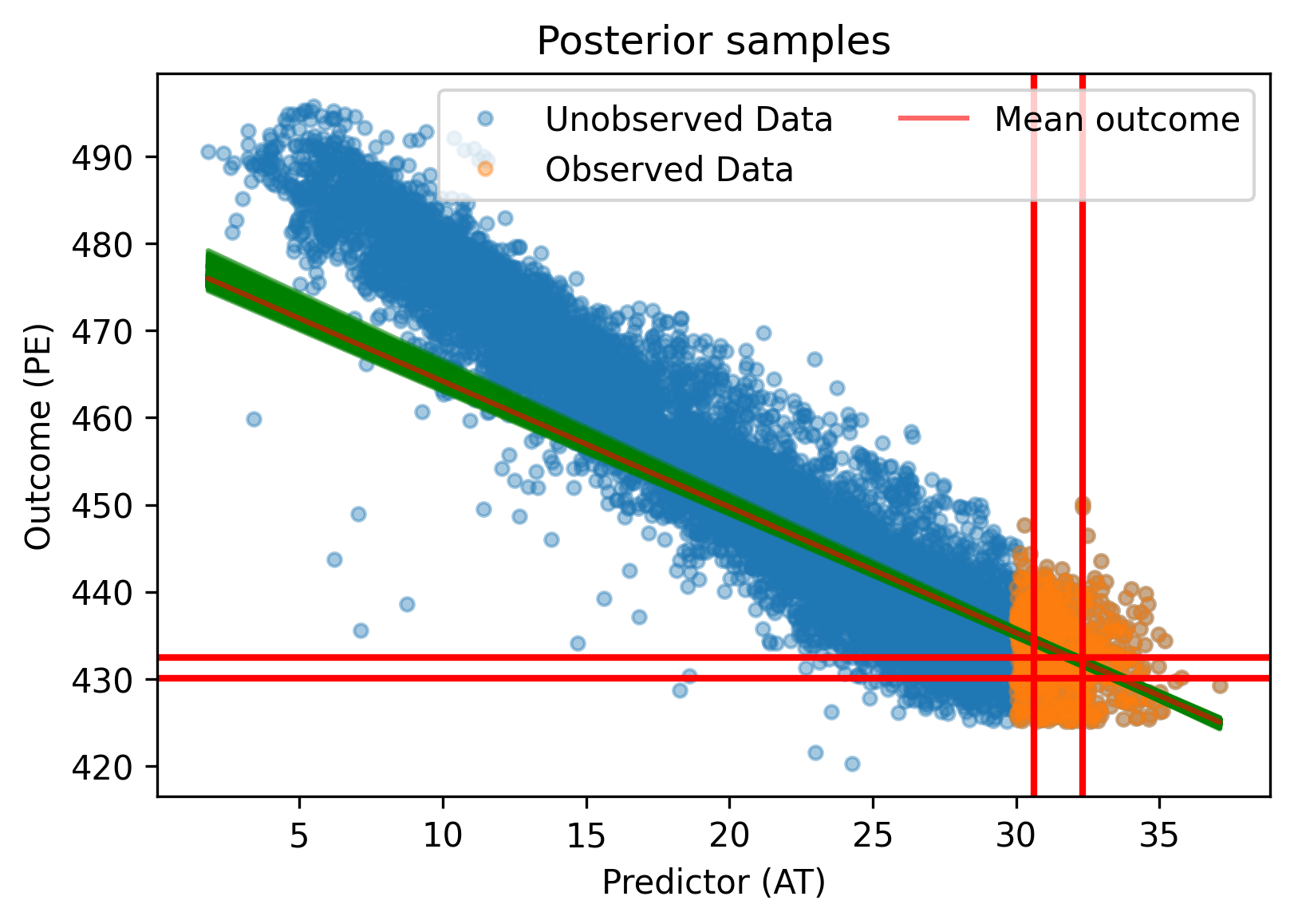}
\caption{\textcolor{black}{Scatterplot of $AT$ and $PE$. The yellow points denote the observed and the blue points the unobserved data. The posterior samples of the linear model with the rules are in green and the mean linear model in dark red. The light red vertical and horizontal lines denote the levels in which the rules are applied ($AT_{low}, AT_{hi}, PE_{low}, PE_{hi}$).}}
\label{fig:CCPP_rules1}
\end{figure}

\textcolor{black}{We repeat the analysis in the previous section with the rule base. The results can be seen in Figure \ref{fig:CCPP_rules1}. The slope is still not precisely the same as the data trend, but a significant improvement is evident. In addition, we can observe a considerable decrease in the posterior uncertainty.}

\textcolor{black}{We compare the metrics for both models in Table \ref{CCPP_metrics}. The model with the rules performs significantly better with all metric values decreasing to less than half the corresponding values for the original model.}

\textcolor{black}{\subsubsection{Remarks}}
\textcolor{black}{This application was indicative of an increase in performance using the rule-based Bayesian variation of a model in a real world dataset example. The rules came from the literature and due to the nature of the statistical model that was used (multivariate linear regression), there was no direct way to include this information (e.g. by using informative priors) without the use of a rule base.}

\textcolor{black}{\subsection{Carbon mono-oxide (CO) emissions from gas turbines}}

\textcolor{black}{For the last application, we aim to predict the $CO$ emission levels of a gas turbine. We derive the data from \citep{kaya2019predicting}, and we fit, once again, a multivariate linear regression model. The sampling scheme is similar to the one from the previous application in Section \ref{subs:CCPP}. }

\textcolor{black}{\subsubsection{Data}\label{sec:gas_em_data}}

\begin{figure}
  \includegraphics[width=\columnwidth]{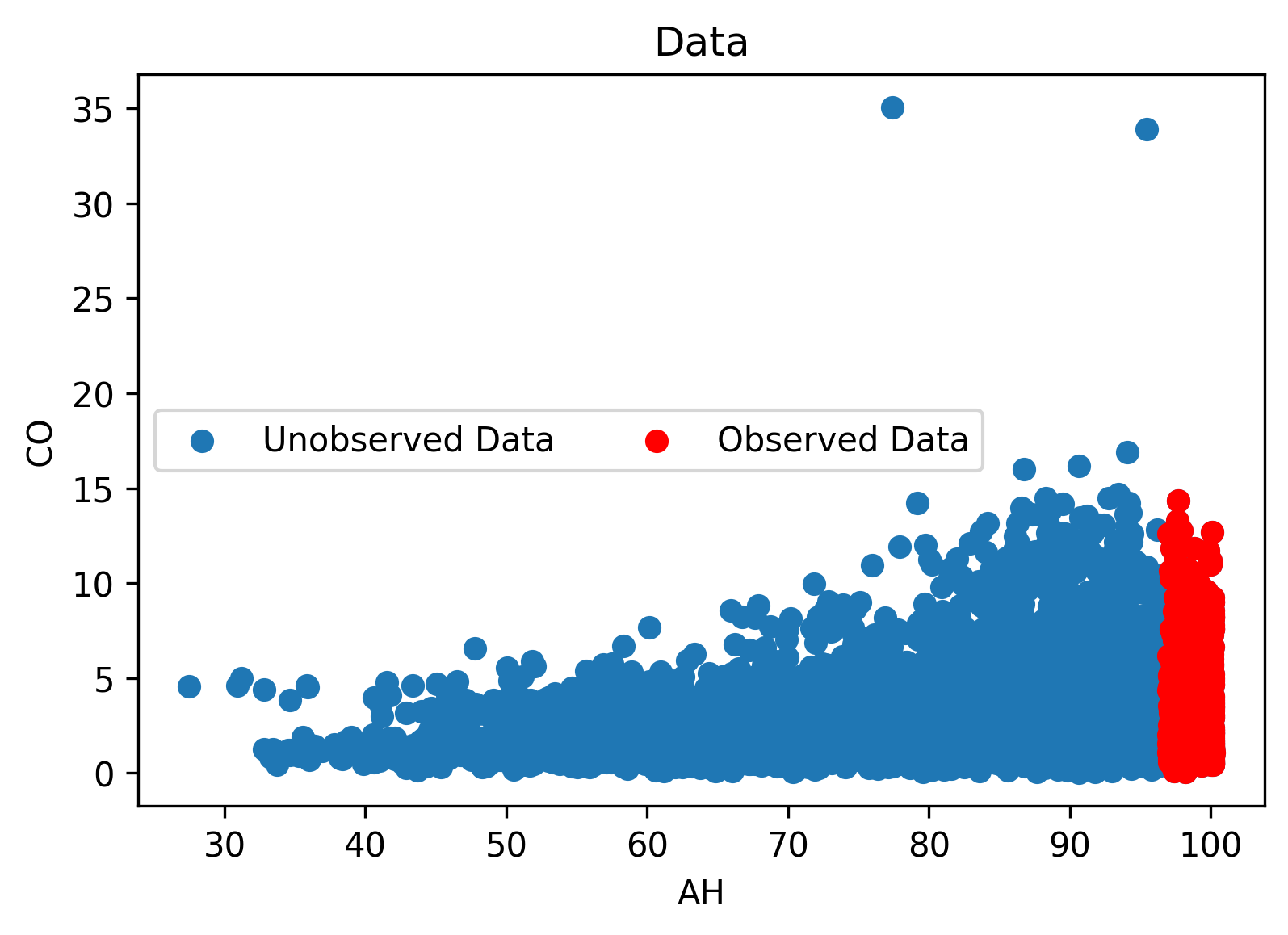}
\caption{\textcolor{black}{Scatterplot of ambient humidity ($AH$) and carbon monoxide emissions ($CO$). The red points denote the observed data (which are used for the model fitting) and the blue points the unobserved data (which are used for the model evaluation).}}
\label{fig:gas_em_data}
\end{figure}

\textcolor{black}{We use the dataset from \citep{kaya2019predicting} that correspond to the year $2013$. It consists of various features and two outputs: the $CO$ and $NOx$ emission levels. We select some of the features to avoid strong correlations among the inputs that effect the efficiency of the sampling algorithm; specifically the ambient temperature $AT$, the ambient humidity $AH$, the air filter difference pressure $AFDP$ and the gas turbine exhaust pressure $GTEP$ and focus only on the $CO$ emissions as output. We limit the training set to the values where the $AH$ is over $95\%$ as shown in Figure \ref{fig:gas_em_data}, replicating a condition where collection of data occurs only during days with very high humidity. This leaves $547$ data points from the original $7152$. The rest of the data is considered the testing set.}

\textcolor{black}{\subsubsection{Multivariate linear regression}}

\textcolor{black}{We construct a multivariate linear regression model with parameters the coefficients of the features mentioned in the previous subsection. The model takes the form:}
\begin{multline*}
    \textcolor{black}{CO = AT_{co}*AT + AH_{co}*AH + AFDP_{co}*AFDP +}\\
    \textcolor{black}{GTEP_{co}*GTEP + b + \epsilon,}
\end{multline*}
\textcolor{black}{where $CO$ are the carbon monoxide emission levels, $AT$, $AH$, $AFDP$ and $GTEP$ are as explained in Section \ref{sec:gas_em_data}, $AT_{co}$, $AH_{co}$, $AFDP_{co}$ and $GTEP_{co}$ are the corresponding coefficients, $b$ is the intercept and $\epsilon$ is Gaussian error with:}
\begin{equation*}
    \textcolor{black}{\epsilon \sim \mathcal{N}(0,1^2).}
\end{equation*}

\textcolor{black}{We choose wide Gaussian distributions for the parameter priors:}
\begin{align*}
    \textcolor{black}{AT_{co}, AH_{co}, AFDP_{co}, GTEP_{co}}  &\textcolor{black}{\sim \mathcal{N}(0, 10^2),} \\
    \textcolor{black}{b} &\textcolor{black}{\sim \mathcal{N}(0, 16^2).}
\end{align*}

\begin{figure}
  \includegraphics[width=\columnwidth]{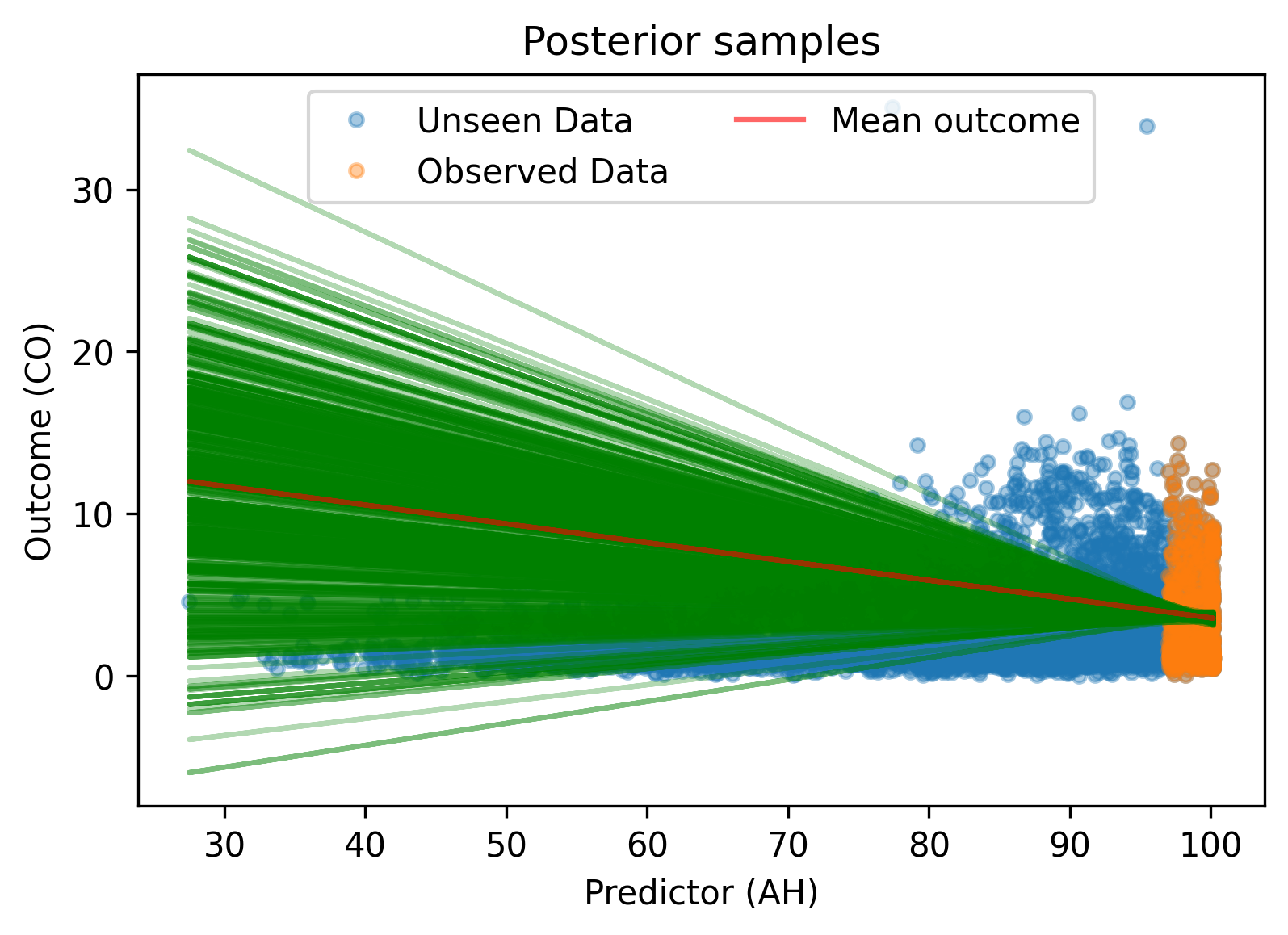}
\caption{\textcolor{black}{Scatterplot of $AH$ and $CO$. The yellow points denote the observed and the blue points the unobserved data. The posterior samples of the linear model without the rules are in green and mean linear model is in dark red.}}
\label{fig:gas_em_post_norules}
\end{figure}

\textcolor{black}{The results are shown in Figure \ref{fig:gas_em_post_norules}, where we show the scatterplot of $AH$ and $CO$, along with the projection of the posterior samples. It is obvious that the posterior uncertainty increases drastically away from the training data, while there is no clear trend. Evaluation metrics are included in Table \ref{emissions_metrics}.}

\begin{table}[h]
\caption{\textcolor{black}{Evaluation metrics for the different models.}}
\centering
\begin{tabular}{|c|c|c|c|}
\hline

\textcolor{black}{Metric/Model} & \textcolor{black}{Without rules}   & \textcolor{black}{With $2$ rules}            & \textcolor{black}{With $3$ rules}\\ \hline
        \textcolor{black}{RMSE}              & \textcolor{black}{\num{1.137}}     & \textcolor{black}{\num{0.741}}     & \textcolor{black}{\num{0.618}}\\ \hline
        \textcolor{black}{MSE}              & \textcolor{black}{\num{1.294}}     & \textcolor{black}{\num{0.550}}     & \textcolor{black}{\num{0.382}}\\ \hline
        \textcolor{black}{MAE}              & \textcolor{black}{\num{0.960}}     & \textcolor{black}{\num{0.550}}     & \textcolor{black}{\num{0.447}}\\ \hline
\end{tabular}
\label{emissions_metrics}
\end{table}

\textcolor{black}{\subsubsection{Rule-based Multivariate linear regression}\label{sec:rule_ex5}}
\textcolor{black}{In order to derive an appropriate rule base for this model, we asked a power plant expert, who came up with the following statement: \textit{$CO$ could be linked with ambient humidity and temperature of the air. If humidity is high, one can expect higher $CO$ emissions as high-water content in the oxidant (air) increase the ignition delay}.}

\textcolor{black}{In mathematical form, this can be written as:}
\begin{align*}
    \textcolor{black}{R_1}&\textcolor{black}{: \text{if} \quad AH \leq AH_{low},\quad \text{then} \quad CO \leq CO_{mid}}\\
    \textcolor{black}{R_2}&\textcolor{black}{: \text{if} \quad AH \geq AH_{hi},\quad \text{then} \quad CO \geq CO_{mid},}
\end{align*}
\textcolor{black}{where $AH_{low}$ is the $25$th percentile, $AH_{hi}$ is the $75$th percentile and $CO_{mid}$ is the $50$th percentile of the observed data.}

\begin{figure}
  \includegraphics[width=\columnwidth]{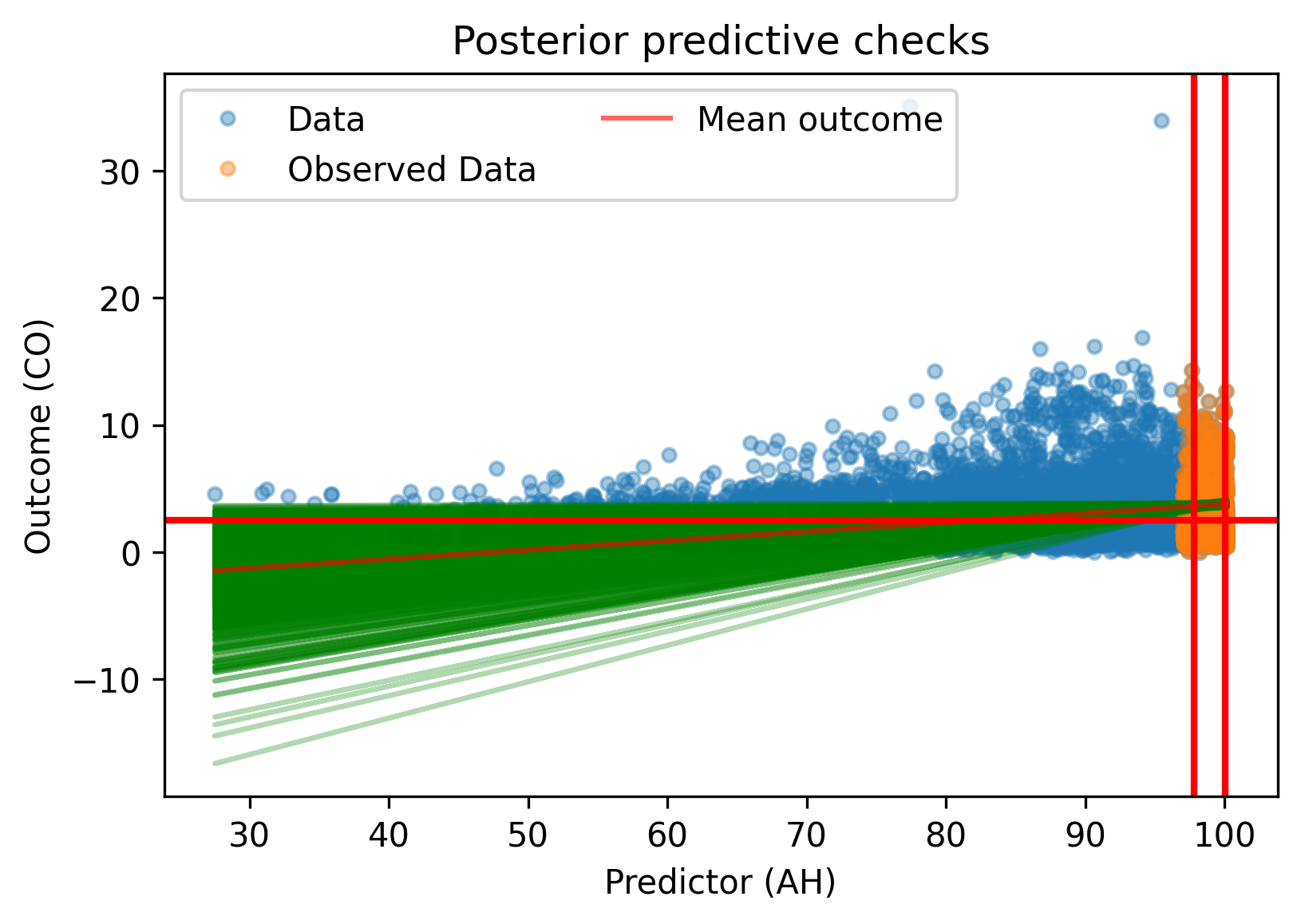}
\caption{\textcolor{black}{Scatterplot of $AH$ and $CO$. The yellow points denote the observed and the blue points the unobserved data. The posterior samples of the linear model with the first set of rules are in green and mean linear model is in dark red. The light red vertical and horizontal lines denote the levels in which the rules are applied ($AH_{low}, AH_{hi}, CO_{mid}$).}}
\label{fig:gas_em_post_rules1}
\end{figure}

\textcolor{black}{We repeated the analysis, including the above rule base. The results can be seen in Figure \ref{fig:gas_em_post_rules1}. There is a very significant decrease in the posterior uncertainty, as well as a clear positive trend.}

\textcolor{black}{Since the rule for small $AH$ values imposes only an upper boundary for the $CO$ values, there is no significant difference for the uncertainty regarding the region with small $AH$ and small $CO$ values. Because of this, the posterior samples favor very low, and even negative regions for the $CO$ emissions, which is not physically possible. This leads us to examine the model with an additional rule, that restricts the model to only positive $CO$ emission levels. This can be incorporated into the model as:}
\begin{align*}
    \textcolor{black}{R_3}&\textcolor{black}{: \text{if} \quad AH \in \mathbb{R},\quad \text{then} \quad CO \geq CO_{low},}
\end{align*}
\textcolor{black}{where $CO_{low} = 0$.}

\begin{figure}
  \includegraphics[width=\columnwidth]{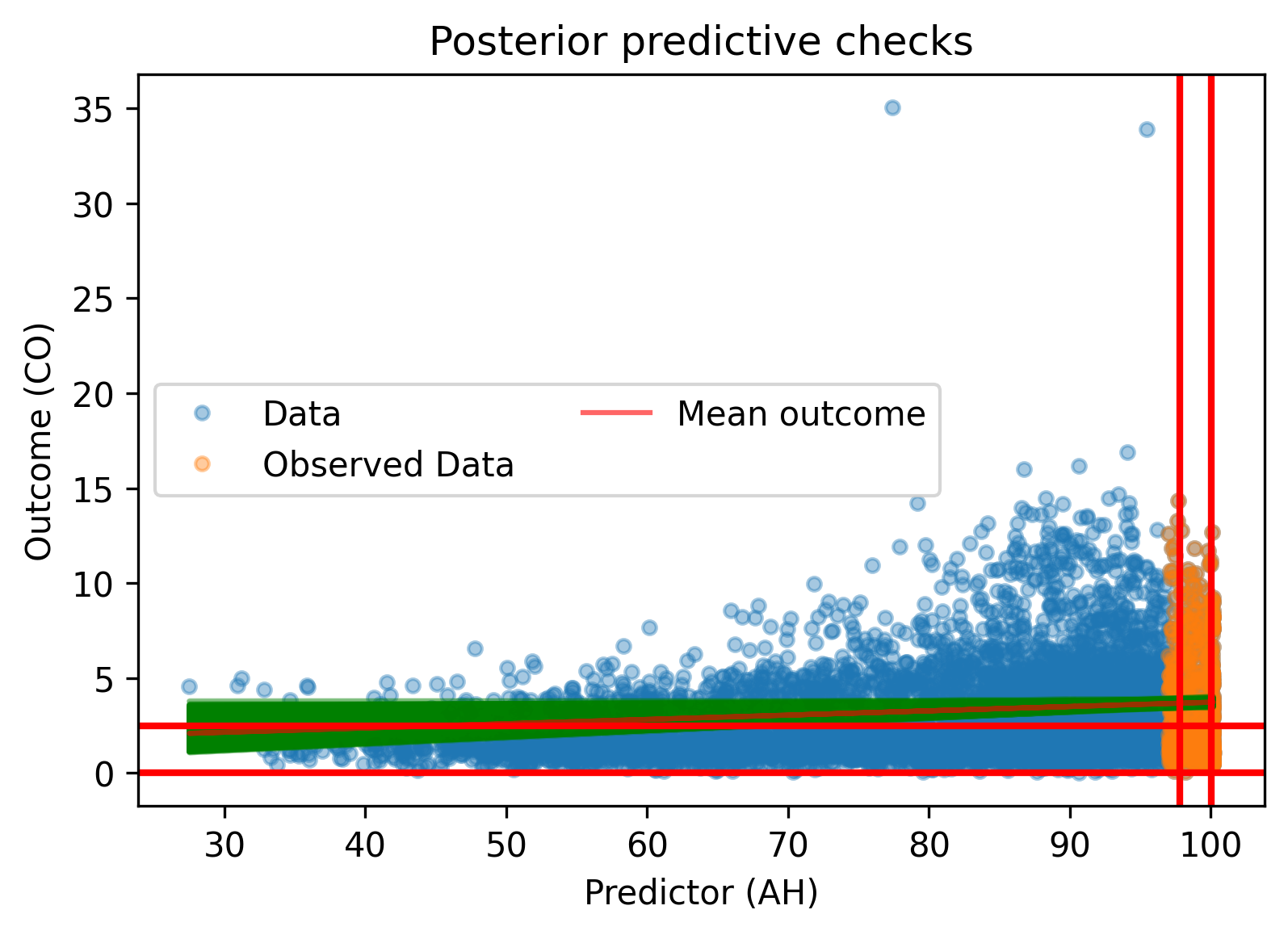}
\caption{\textcolor{black}{Scatterplot of $AH$ and $CO$. The yellow points denote the observed and the blue points the unobserved data. The posterior samples of the linear model with the second set of rules are in green and mean linear model is in dark red. The light red vertical and horizontal lines denote the levels in which the rules are applied ($AH_{low}, AH_{hi}, CO_{low}, CO_{mid}$).}}
\label{fig:gas_em_post_rules2}
\end{figure}

\textcolor{black}{The results including all three rules are shown in Figure \ref{fig:gas_em_post_rules2}. The uncertainty has, again, decreased significantly and the mean prediction still indicates a positive trend, but this time limiting $CO$ to only positive values.}

\textcolor{black}{Analysing the metrics in Table \ref{emissions_metrics}, we observe a substantial improvement between the model without rules and the ones where the rule-based method was performed. There was a smaller increase in performance between the model with the two rules and the one with the three rules. The more significant improvement that came from the inclusion of the third rule was the very substantial decrease in uncertainty.}

\textcolor{black}{\subsubsection{Remarks}}

\textcolor{black}{The key element from this application was the combination of rules that come from an actual field expert and also directly from the physics of the system. In addition, it was important to notice that, except from the best fit improvement, additional rules might have a substantial effect in the system uncertainty.}

\section{Discussion}\label{discussion}

The major issue that needs to be addressed in order to extend the method to increasingly realistic applications is the computational complexity. As the rule base increases, the structure becomes more computationally expensive. This is because MCMC techniques and optimisation methods with multiple calls are required, while the rule structure needs to be included in each algorithmic iteration. Some approaches that can address this issue include specific variations of machine learning algorithms that can account for expensive computations (e.g.~sparse approximations) and sub-sampling of the data.

Another complication is related to the connection of the posterior distribution, as it is modified by the rule base, and the sampling algorithms. For typical sampling techniques, the starting point plays an important role, since, if it is assigned a `heavy' penalty by the rules, it can potentially affect the efficiency of the algorithm. On a similar note, the rule structure can render the posterior multi-modal, an issue where most conventional algorithms struggle. A more robust sampling technique that can examine large regions, by sampling specific points in the region and assigning an initial regional penalty, could address this issue and decrease substantially the computational time required for the solution of rule-based Bayesian regression problems.

As shown in the applications of Section \ref{Applications}, the methodology can easily be incorporated into probabilistic programming frameworks such as Stan \citep{stan_cite}, PyMC3 \citep{Salvatier2016} and TensorFlow Probability \citep{abadi2016tensorflow}. Most traditional probabilistic programming tools are taking advantage of a directed acyclic graph structure and the rule-based section of the method can easily be depicted as an extra vertex, usually in the form of a custom likelihood.

It is important to acknowledge that, as with informative priors in a standard Bayesian context, the intuition of the expert, which dictates the rules, plays a critical role to the performance of the model and the validity of the predictions. Ineffective or poor-quality rules can lead to misleading results.

\section{Conclusion}\label{conclusion}

We have introduced a new formalism that aims to merge the main advantages of Bayesian inference and rule-based systems: incorporating domain expertise into the model through the latter and using the former for uncertainty quantification.
We presented the general framework for the rule-based Bayesian regression and we treated the GP regression as a special case.
We used our methodology in \textcolor{black}{five} applications, adopting different statistical models: in the first, data were derived from a linear model and we used a linear regression model, in the second, we derived data from a one-dimensional velocity advection equation and we used third-degree B-splines, in the third, we used GPs in order to emulate a two-dimensional advection-diffusion equation,  \textcolor{black}{and, finally, in the fourth and fifth, we used multivariate linear regression models for the electrical output of a combined cycle power plant and the $CO$ emissions from a gas turbine respectively}.

We also demonstrated variations that display the flexibility of the method and show how it can be used in order to model different levels of confidence regarding inputs from expert intuition, including changes in the value of relevant parameters and the use of hyperparameters.

Future research should be focused in applying the methodology to \textcolor{black}{more complex} real data applications, where the new challenges mentioned in Section \ref{discussion} might \textcolor{black}{be more prominent}, such as computational issues and difficulty of articulating domain expertise into a rule base. \textcolor{black}{Finally, part of future plans includes the creation of a framework where competing rules can be compared and assessed, by expanding \citep{pan2018marginal}.}

\begin{acknowledgements}
This work was supported by Wave 1 of The UKRI Strategic Priorities Fund under the EPSRC Grant EP/T001569/1, particularly the \emph{Digital Twins for Complex Engineering Systems} theme within that grant and The Alan Turing Institute. IP acknowledges funding from the Imperial College Research Fellowship scheme. We acknowledge Dr. Daya Shankar Pandey at University of Huddersfield, UK, who is a power plant expert and helped with the rule elicitation in Section \ref{sec:rule_ex5}.
\end{acknowledgements}

\section*{Conflict of interest}

The authors declare that they have no conflict of interest.

\bibliography{bibliography}

\begin{thebibliography}{}

\bibitem[Abadi et~al., 2016]{abadi2016tensorflow}
Abadi, M., Barham, P., Chen, J., Chen, Z., Davis, A., Dean, J., Devin, M.,
  Ghemawat, S., Irving, G., Isard, M., et~al. (2016).
\newblock Tensorflow: A system for large-scale machine learning.
\newblock In {\em 12th $\{$USENIX$\}$ Symposium on Operating Systems Design and
  Implementation ($\{$OSDI$\}$ 16)}, pages 265--283.

\bibitem[Bar-Sinai et~al., 2019]{bar2019learning}
Bar-Sinai, Y., Hoyer, S., Hickey, J., and Brenner, M.~P. (2019).
\newblock Learning data-driven discretizations for partial differential
  equations.
\newblock {\em Proceedings of the National Academy of Sciences},
  116(31):15344--15349.

\bibitem[Bishop, 2006]{bishop2006pattern}
Bishop, C.~M. (2006).
\newblock {\em Pattern Recognition and Machine Learning}.
\newblock springer.

\bibitem[Breiman et~al., 1984]{breiman1984classification}
Breiman, L., Friedman, J., Stone, C.~J., and Olshen, R.~A. (1984).
\newblock {\em Classification and Regression Trees}.
\newblock CRC press.

\bibitem[Ching and Chen, 2007]{ching2007transitional}
Ching, J. and Chen, Y.-C. (2007).
\newblock {Transitional Markov chain Monte Carlo method for Bayesian model
  updating, model class selection, and model averaging}.
\newblock {\em Journal of Engineering Mechanics}, 133(7):816--832.

\bibitem[Chipman et~al., 2010]{chipman2010bart}
Chipman, H.~A., George, E.~I., McCulloch, R.~E., et~al. (2010).
\newblock {BART: Bayesian additive regression trees}.
\newblock {\em The Annals of Applied Statistics}, 4(1):266--298.

\bibitem[de~Boor, 1978]{de1978practical}
de~Boor, C. (1978).
\newblock {\em A Practical Guide to Spline}, volume Volume 27.
\newblock Springer (New York, NY [ua]).

\bibitem[Gelman et~al., 2013]{gelman2013bayesian}
Gelman, A., Carlin, J.~B., Stern, H.~S., Dunson, D.~B., Vehtari, A., and Rubin,
  D.~B. (2013).
\newblock {\em {Bayesian Data Analysis}}.
\newblock CRC press.

\bibitem[Gonz{\'a}lez-D{\'\i}az et~al., 2017]{gonzalez2017effect}
Gonz{\'a}lez-D{\'\i}az, A., Alcar{\'a}z-Calder{\'o}n, A.~M.,
  Gonz{\'a}lez-D{\'\i}az, M.~O., M{\'e}ndez-Aranda, {\'A}., Lucquiaud, M., and
  Gonz{\'a}lez-Santal{\'o}, J.~M. (2017).
\newblock Effect of the ambient conditions on gas turbine combined cycle power
  plants with post-combustion {CO}2 capture.
\newblock {\em Energy}, 134:221--233.

\bibitem[Hastings, 1970]{hastings1970monte}
Hastings, W.~K. (1970).
\newblock {M}onte {C}arlo sampling methods using {M}arkov chains and their
  applications.
\newblock {\em Biometrika}, 57(1):97--109.

\bibitem[Hoyer and Zhuang, 2020]{Hoyer2020data}
Hoyer, S. and Zhuang, J. (2020).
\newblock {\em {Data driven discretizations for solving 2D PDEs}}.
\newblock \url{https://github.com/google-research/data-driven-pdes}.

\bibitem[Kaya et~al., 2019]{kaya2019predicting}
Kaya, H., T{\"u}fekc{\.i}, P., and Uzun, E. (2019).
\newblock Predicting {CO} and {NOx} emissions from gas turbines: novel data and
  a benchmark {PEMS}.
\newblock {\em Turkish Journal of Electrical Engineering \& Computer Sciences},
  27(6):4783--4796.

\bibitem[Kharratzadeh, 2017]{splines_stan}
Kharratzadeh, M. (2017).
\newblock {Splines in Stan}.
\newblock
  \url{https://github.com/milkha/Splines_in_Stan/blob/master/splines_in_stan.pdf}.

\bibitem[Lakshminarayanan et~al., 2016]{lakshminarayanan2016mondrian}
Lakshminarayanan, B., Roy, D.~M., and Teh, Y.~W. (2016).
\newblock Mondrian forests for large-scale regression when uncertainty matters.
\newblock In {\em Artificial Intelligence and Statistics}, pages 1478--1487.

\bibitem[Lundberg et~al., 2020]{lundberg2020local}
Lundberg, S.~M., Erion, G., Chen, H., DeGrave, A., Prutkin, J.~M., Nair, B.,
  Katz, R., Himmelfarb, J., Bansal, N., and Lee, S.-I. (2020).
\newblock From local explanations to global understanding with explainable {AI}
  for trees.
\newblock {\em Nature Machine Intelligence}, 2(1):2522--5839.

\bibitem[Minson et~al., 2013]{minson2013bayesian}
Minson, S., Simons, M., and Beck, J. (2013).
\newblock {Bayesian inversion for finite fault earthquake source models
  I—Theory and algorithm}.
\newblock {\em Geophysical Journal International}, 194(3):1701--1726.

\bibitem[Nelder and Mead, 1965]{nelder1965simplex}
Nelder, J.~A. and Mead, R. (1965).
\newblock A simplex method for function minimization.
\newblock {\em The Computer Journal}, 7(4):308--313.

\bibitem[O’Hagan, 2019]{o2019expert}
O’Hagan, A. (2019).
\newblock Expert knowledge elicitation: subjective but scientific.
\newblock {\em The American Statistician}, 73(sup1):69--81.

\bibitem[Pan and Bester, 2017]{pan2017fuzzy}
Pan, I. and Bester, D. (2017).
\newblock {Fuzzy Bayesian learning}.
\newblock {\em IEEE Transactions on Fuzzy Systems}, 26(3):1719--1731.

\bibitem[Pan and Bester, 2018]{pan2018marginal}
Pan, I. and Bester, D. (2018).
\newblock {Marginal likelihood based model comparison in Fuzzy Bayesian
  Learning}.
\newblock {\em IEEE Transactions on Emerging Topics in Computational
  Intelligence}, 4(6):794--799.

\bibitem[Rasmussen, 2003]{rasmussen2003gaussian}
Rasmussen, C.~E. (2003).
\newblock Gaussian processes in machine learning.
\newblock In {\em Summer School on Machine Learning}, pages 63--71. Springer.

\bibitem[Rochford, 2017]{splines_pymc3}
Rochford, A. (2017).
\newblock {A {PyMC}3 port of Splines in Stan}.
\newblock
  \url{https://gist.github.com/AustinRochford/d640a240af12f6869a7b9b592485ca15}.

\bibitem[Salvatier et~al., 2016]{Salvatier2016}
Salvatier, J., Wiecki, T.~V., and Fonnesbeck, C. (2016).
\newblock Probabilistic programming in python using {PyMC}3.
\newblock {\em {PeerJ} Computer Science}, 2:e55.

\bibitem[{Stan Development Team}, 2019]{stan_cite}
{Stan Development Team} (2019).
\newblock {RStan}: the {R} interface to {Stan}.
\newblock R package version 2.19.1.

\bibitem[T{\"u}fekci, 2014]{tufekci2014prediction}
T{\"u}fekci, P. (2014).
\newblock Prediction of full load electrical power output of a base load
  operated combined cycle power plant using machine learning methods.
\newblock {\em International Journal of Electrical Power \& Energy Systems},
  60:126--140.

\end{thebibliography}

\end{document}